%% file: NAACL_main.tex
\documentclass[11pt]{article}

\usepackage[preprint]{acl}

\usepackage{times}
\usepackage{latexsym}

\usepackage[T1]{fontenc}

\usepackage[utf8]{inputenc}

\usepackage{microtype}

\usepackage{inconsolata}

\usepackage{graphicx}

\usepackage{booktabs}
\usepackage{multirow}

\title{FOLD: Position-Free Far Memory with Local Order for Long-Context Speculative Decoding}

\author{
  \textbf{Shuxuan Li}$^{1,2}$ \quad
  \textbf{Jialiang Kang}$^{2}$ \quad
  \textbf{Yanghao Zhou}$^{2}$ \\
  \textbf{Ziye Zhu}$^{2,3}$ \quad
  \textbf{Zhilin Zhao}$^{1}$\thanks{Corresponding authors.
    Code is available at \url{https://github.com/NuyoahNasuS/FOLD}.} \quad
  \textbf{Xiao Yang}$^{2}$\footnotemark[1] \\
  $^{1}$Sun Yat-sen University \quad
  $^{2}$Canva Research \quad
  $^{3}$Peking University
}

\begin{document}
\maketitle

\begin{abstract}
Speculative decoding accelerates LLM generation through lightweight drafting and parallel verification. It is particularly promising for long inputs, where KV-cache access costs can be amortized over multiple accepted tokens. Yet most drafters are trained on short sequences and degrade on long inputs, while long-context training is costly.
Existing training-free approaches restrict the context window, rescale positions, or augment recent context with retrieved distant entries.
Drafters, however, read target hidden states that already encode contextual and positional information, and their lightweight architecture makes full-history attention affordable.
We propose FOLD, a training-free method that preserves exact local positions and treats all distant entries as far memory at a shared position.
To counter attention dilution as the memory grows, we sharpen attention within it to emphasize relevant entries while separately controlling its total weight relative to the local window.
Across six target--drafter pairs and diverse benchmarks of up to 256K tokens, FOLD predominantly delivers the highest speedup, with the best case more than doubling the speedup over autoregressive decoding, from 2.08$\times$ to 4.22$\times$.
\end{abstract}




\input{sections/01Intro}
\input{sections/02RelatedWorks}
\input{sections/03Method}
\input{sections/04Experiment}
\input{sections/05Conclusion}

\section*{Limitations}

Acceptance length, and hence the speedup, depends strongly on the target model and the task, so the gains of FOLD vary across models and tasks.
Because FOLD changes only how the drafter reads its context, it cannot address degradation caused by the target itself, whose features and predictions change with input length, which may limit its gains on models such as Gemma-4-12B.
We focus on block-parallel drafters and do not evaluate autoregressive drafters such as EAGLE-3~\cite{li2025eagle3}, since block-parallel drafters have been reported to achieve substantially higher speedups~\cite{chen2026dflash}.
FOLD also keeps an extra, unrotated copy of the drafter's keys and values, whose size grows linearly with context length.
This overhead is small for lightweight drafters, but it can become significant under tight memory budgets or high concurrency.
The window size $W$ and concentration factor $\alpha$ are set manually and remain fixed during decoding.
Efficient adaptation of these parameters during inference is left for future work.




\bibliography{custom}

\input{sections/Appendix}




\end{document}

%% file: sections/01Intro.tex
\section{Introduction}
\label{sec:Intro}

\begin{figure}[t]
  \centering
  \includegraphics[width=\columnwidth]{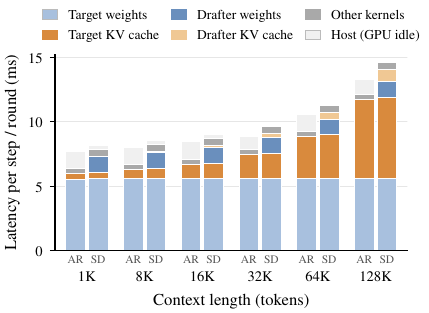}
  \caption{\textbf{Latency breakdown across context lengths} for Llama-3.1-8B (vLLM, one H100 GPU, batch size~1), for an AR step and a DFlash SD round with block size 10. From 1K to 128K tokens, target KV-cache attention grows from 5\% to 46\% of the AR step, while drafter KV-cache attention stays below 7\% of the SD round.}
  \label{fig:decode_breakdown}
\end{figure}

Large language models (LLMs) are increasingly deployed in long-context applications, such as long-document understanding, repository-scale coding, and agentic workflows, where inputs can span hundreds of thousands of tokens~\cite{zhu2026tracelab,yang2026longspec}.
At these lengths, attention over the KV cache becomes a major cost of decoding. For Llama-3.1-8B at 128K tokens, it takes 46\% of an autoregressive (AR) decoding step, more than the weight matrix multiplications (Figure~\ref{fig:decode_breakdown}).
Speculative decoding (SD)~\cite{leviathan2023fast,chen2023accelerating} accelerates decoding without changing the output distribution, as a lightweight drafter proposes tokens that the target model verifies in parallel. This amortizes weight and KV cache reads over accepted tokens and pays off more as the context grows~\cite{sadhukhan2025magicdec}.
Block-parallel drafters such as DFlash~\cite{chen2026dflash} and DSpark~\cite{cheng2026dspark} further reduce drafting cost by proposing a block of tokens at once.

\begin{figure}[t]
  \centering
  \includegraphics[width=\columnwidth]{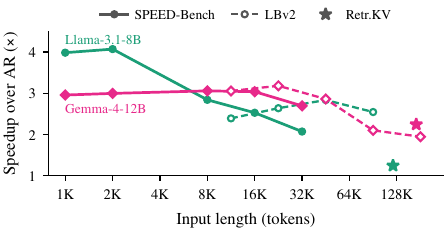}
  \caption{\textbf{Long-context degradation of DFlash drafters.} Speedup over AR decoding for Llama-3.1-8B and Gemma-4-12B on SPEED-Bench (1K--32K tokens)~\cite{abramovich2026speedbench}, LongBench~v2~\cite{bai2025longbenchv2} (grouped by input length), and Retr.KV~\cite{zhang2024infinitebench}.}
  \label{fig:length_speedup_down}
\end{figure}

Cheap drafting, however, helps only if the drafts are accepted.
Most existing drafters are trained on short sequences, for example 3K tokens for the Llama-3.1-8B DFlash drafter~\cite{chen2026dflash}, and degrade substantially on longer inputs.
As Figure~\ref{fig:length_speedup_down} shows, the speedup of the Llama-3.1-8B DFlash drafter nearly halves within 32K tokens, and that of the Gemma-4-12B drafter falls from 3.18$\times$ to 1.94$\times$ as LongBench~v2 inputs grow beyond 128K.
Long-context fine-tuning and dedicated length-generalization training can mitigate this degradation~\cite{chen2026dflash,yang2026longspec}, but require additional data and training for each target--drafter pair, motivating training-free adaptation of existing drafters.

Prior work attributes much of this degradation to relative positions beyond the training range~\cite{chen2026dflash,yang2026longspec}.
Training-free remedies, developed mostly for language models and partly adopted in SD, handle these positions differently.
StreamingLLM~\cite{xiao2024streamingllm} retains attention sinks and a recent window but discards the remaining history, a strategy also adopted for self-speculative drafting in MagicDec~\cite{sadhukhan2025magicdec}.
Position scaling methods such as YaRN~\cite{peng2024yarn}, also applied to block-parallel drafters~\cite{rusanovsky2026lilicorr}, retain the full context but remap all positions.
ReRoPE~\cite{su2023rerope} preserves exact local distances and caps distant ones, with length-dependent scaling applied to all attention logits to counter attention dilution.
InfLLM~\cite{xiao2024infllm} and SpecExtend~\cite{cha2026specextend} augment recent context with retrieved distant blocks, introducing selection and refresh overhead.

To address positional extrapolation, we seek to keep the drafter's effective relative distances within its training range while retaining distant information.
Block-parallel drafters such as DFlash derive their historical keys and values from target hidden states that already encode contextual and positional information.
Our findings suggest that distant states can serve as content-bearing memory at bounded distances, while precise local positions preserve the short-context structure needed for upcoming predictions.
Reading this memory in full is also affordable for the lightweight drafter, whose KV-cache attention accounts for only a small fraction of SD-round latency (Figure~\ref{fig:decode_breakdown}).
Retaining all history thus avoids selection and refresh overhead that can offset the savings from attending to fewer entries and erode SD speedup.

Building on this view, we propose FOLD, a training-free method that preserves exact relative positions within a local window to retain the positional structure learned on short sequences and maps all distant entries to a shared position just beyond it, forming a far memory.
To counter attention dilution as the memory grows, FOLD normalizes the window and the memory separately and uses a concentration factor $\alpha$ to control both how attention is distributed within the memory and how much total weight it receives.
FOLD leaves target-model verification unchanged and recovers the original attention computation when the context fits within the window.
We evaluate FOLD across six target–drafter pairs with inputs of up to 256K tokens. On seven long-context benchmarks, it consistently raises the average speedup over AR decoding of the original drafters, by 9\% to 61\% across Llama-3.1-8B, Qwen3.5-9B, and Gemma-4-12B.
The gains also extend to DSpark drafters and the SGLang inference engine. Our contributions are as follows:

\begin{itemize}
\item We analyze the behavior of long-context drafting, finding that large relative distances can reduce acceptance and that target-derived features encode coarse positional information.
\item We propose FOLD, which retains distant context at a shared position and controls attention dilution through decoupled
normalization.
\item We systematically evaluate and analyze FOLD's performance across drafter designs and input lengths.
\end{itemize}

%% file: sections/02RelatedWorks.tex
\section{Related Work}
\label{sec:related_work}

\subsection{Speculative Decoding for Long Contexts}
\label{sec:rw_speculative_decoding}

Speculative decoding~\cite{leviathan2023fast,chen2023accelerating} accelerates generation as a lightweight drafter proposes tokens that the target model verifies in parallel, without changing the output distribution.
Recent drafters condition on target hidden features and draft either autoregressively~\cite{li2025eagle3} or a whole block at once~\cite{chen2026dflash,huang2026domino,cheng2026dspark}.
Long inputs make SD more valuable, as KV-cache reads are amortized over accepted tokens~\cite{sadhukhan2025magicdec}, yet drafters trained on short sequences lose much of their acceptance length on such inputs~\cite{cha2026specextend,chen2026dflash}.
Existing remedies mostly target autoregressive drafting: self-speculative methods draft with the target model over a sparse KV cache~\cite{sadhukhan2025magicdec,sun2024triforce,yue2026vegas}, while separate drafters are trained for length generalization~\cite{yang2026longspec} or read context selected by target attention~\cite{cha2026specextend}.
Block-parallel drafters have been adapted through long-context fine-tuning~\cite{chen2026dflash}, inference-time RoPE scaling~\cite{rusanovsky2026lilicorr}, or sliding-window attention~\cite{jiang2026hspec}, with evaluations on inputs of up to 32K tokens.
Adapting pretrained block-parallel drafters of different designs to contexts beyond 100K tokens remains underexplored.

\subsection{Context Extension for LLMs}
\label{sec:rw_context_extension}

Training-free context extension either adapts positional encodings, through rotary rescaling~\cite{chen2023extending,peng2024yarn} and relative-position remapping~\cite{su2023rerope,jin2024llm,an2024training,an2025string,tang2026jetlong}, or restricts the attended context, through attention sinks with a recent window~\cite{han2024lminfinite,xiao2024streamingllm} and retrieval of distant blocks~\cite{xiao2024infllm}. Several of these methods keep exact positions locally and coarsen distant ones, by grouping in Self-Extend~\cite{jin2024llm} or by assigning all distant tokens one shared distance in ReRoPE~\cite{su2023rerope} and InfLLM~\cite{xiao2024infllm}, the treatment closest to FOLD.
FOLD applies this separation to feature-conditioned drafting, motivated by the coarse position already encoded in target hidden states (\S\ref{sec:diagnosis}).
Unlike ReRoPE, it normalizes the distant memory separately from the window, and unlike InfLLM, it reads every distant entry, since attending to the full history is cheaper for a drafter than retrieval.

%% file: sections/03Method.tex
\section{FOLD: Far Memory with Local Order}
\label{sec:method}

\subsection{Preliminaries: Block-Parallel Drafting}
\label{sec:prelim}

We consider a target model $\mathcal{T}$ accelerated by a block-parallel drafter $\mathcal{D}$ in the style of DFlash~\cite{chen2026dflash}.
Let $x_{1:t}$ denote the current token prefix, with $x_t$ sampled by $\mathcal{T}$ in the last verification step.
For each $j < t$, hidden states from several layers of $\mathcal{T}$ are concatenated into $h_j$ and projected to a context feature $c_j = \mathrm{Norm}(W_{\mathrm{fc}}\, h_j)$, from which each drafter layer derives a context key $k_j$ and value $v_j$.
The draft block spans positions $p = t, \ldots, t+B-1$, the first holding the embedding of $x_t$ and the others a mask token.

Block positions attend bidirectionally to one another and to all context entries.
Under rotary position embeddings (RoPE)~\cite{su2024roformer}, the logit between block position $p$ and context entry $j$ is $s_{p,j}=q_p^{\top}\mathbf{R}_{p-j}k_j/\sqrt{d}$, where $q_p$ is the query at $p$, $\mathbf{R}_{\Delta}$ is the rotation for distance $\Delta$, and $d$ is the head dimension, so its positional dependence is determined by the relative distance $p-j$.
One drafter forward pass proposes $B-1$ tokens ($B$ for DeepSpec drafters), and $\mathcal{T}$ verifies them in one forward pass, committing the longest accepted prefix together with one token sampled by $\mathcal{T}$.
Following prior work~\cite{li2025eagle3,chen2026dflash}, we measure drafting quality by the acceptance length $\tau$, the average number of tokens committed per round.

Drafters are typically trained on target-generated sequences of only a few thousand tokens.
Unlike a standalone language model, the drafter reads features $c_j$ rather than raw tokens, with each $x_j$ already contextualized within the ordered prefix.
Despite this target-side contextualization, the logit still depends on the relative distance $p-j$, which for distant entries can far exceed the range seen in training, placing them outside the drafter's learned positional regime, where they can disrupt the attention patterns learned on short contexts.

\subsection{Diagnosing Long-Context Drafting}
\label{sec:diagnosis}

Two findings motivate the design of FOLD. The first isolates the effect of large relative distances on drafting, and the second probes the positional order that the context features already carry.

\begin{figure}[t]
  \centering
  \includegraphics[width=\columnwidth]{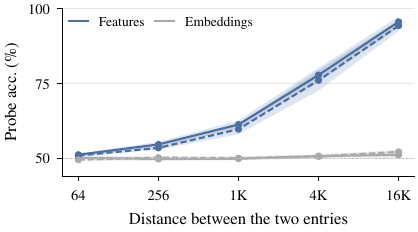}
  \caption{\textbf{Diagnosing long-context drafting on Llama-3.1-8B.}
  Accuracy of a linear probe predicting the order of two entries from the difference of their context features or token embeddings, by separation, on SPEED-Bench 32K (solid) and LBv2 (dashed), with bands of one standard deviation over five document-level folds.}
  \label{fig:diagnosis}
\end{figure}

\paragraph{Finding 1: Large relative distances alone can disrupt drafting.}
To isolate the effect of distance from that of context content, we shift the draft block by 8K--128K positions relative to all committed entries while keeping the content and all within-block distances fixed (Appendix~\ref{app:fixed_state}).
On 2K-token SPEED-Bench~\cite{abramovich2026speedbench} inputs, this intervention reduces the fraction of accepted draft tokens from 0.351 to at most 0.014 for Llama-3.1-8B and from 0.241 to at most 0.076 for Gemma-4-12B.
Thus, increasing relative distances alone can sharply reduce draft acceptance.

\paragraph{Finding 2: Context features carry coarse position.}
A linear probe trained on $c_i-c_j$ predicts the order of two Llama-3.1-8B context entries with 94--96\% accuracy when they are separated by at least 16K tokens, but performs near chance (51\%) at separations of 64--256 tokens, while the corresponding probe on token embeddings remains at chance (Figure~\ref{fig:diagnosis}).
When $\mathcal{T}$ processes the same entries in a shuffled order, the probe recovers this processing order with 84--86\% accuracy, but the original order with only 52\%.
This suggests that coarse positional information is written into $c_j$ through $\mathcal{T}$'s position-aware computation~\cite{haviv2022transformer}.

Together, these findings motivate retaining precise positional order near the draft block while reading distant entries through their contextualized features, without imposing extrapolated distances.

\subsection{Far Memory with Local Order}
\label{sec:fold}

\begin{figure*}[t]
  \centering
  \includegraphics[width=\textwidth]{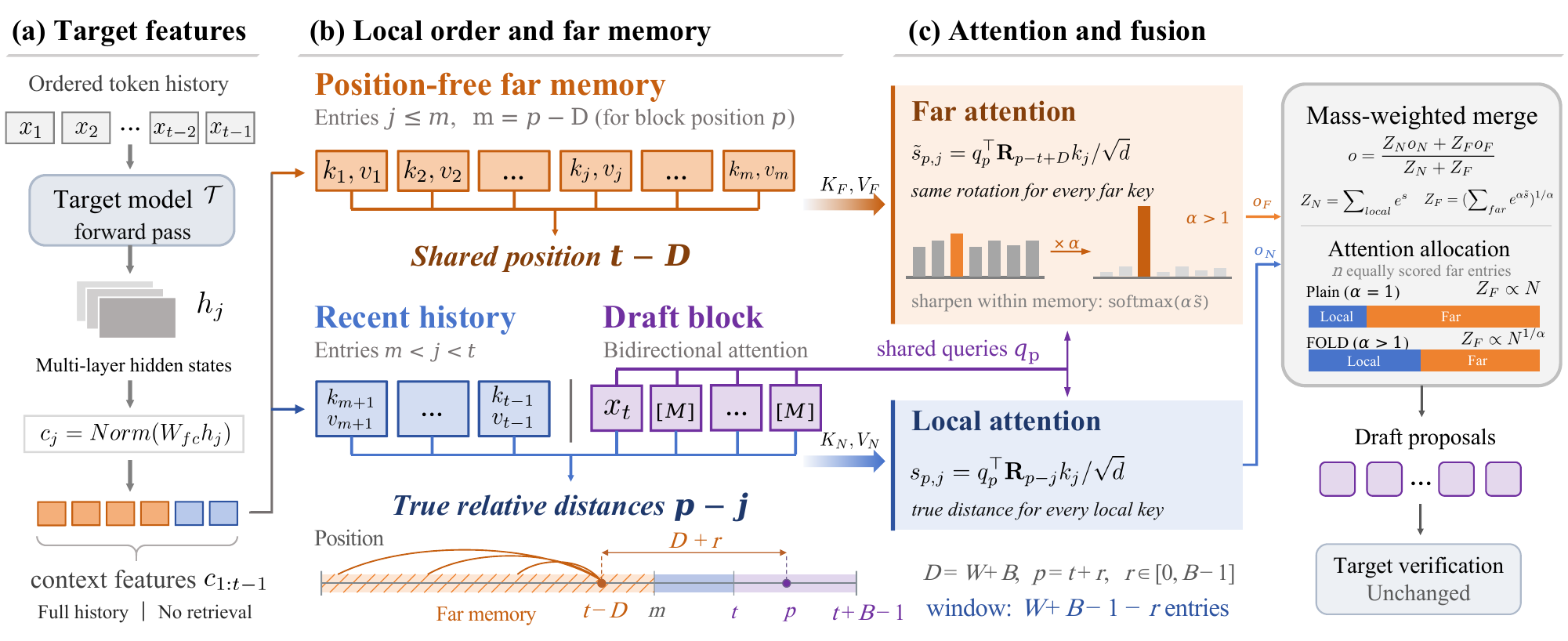}
  \caption{\textbf{Overview of FOLD.}
  (a) Target hidden states of the ordered prefix are projected into context features $c_j$.
  (b) The draft block and recent entries keep their true relative distances, and all earlier entries share the virtual position $t-D$.
  (c) The window and the far memory are attended separately, and $\alpha$ sharpens the memory read while its mass grows as $N^{1/\alpha}$ rather than $N$ for $N$ equally scored entries. The outputs are merged by their masses.}
  \label{fig:fold}
\end{figure*}

Guided by these findings, FOLD splits the drafter's context into a recent window and a far memory (Figure~\ref{fig:fold}).
This split reflects the different roles of recent and distant context during generation.
Recent context supports the syntax, style, and discourse trajectory of the local continuation, where precise token order matters.
Distant context instead serves primarily as a content-addressable memory for retrieving earlier facts and evidence, where preserving information matters more than exact relative distances.
FOLD changes only how the drafter attends to its context, leaving the drafter's weights, the target model, and verification unchanged.
It has two inference-time hyperparameters, the window size $W$ and the concentration factor $\alpha$.

\paragraph{Position-free far memory.}
Let $W$ denote the window size and set $D=W+B$.
For each block position $p$, the local window $\mathcal{N}_p$ contains all block positions and context entries satisfying $p-j<D$.
These entries retain their true relative distances, and $\mathcal{N}_p$ includes the most recent $\min(W,t-1)$ context entries.
All remaining context entries, $\mathcal{F}_p=\{j:p-j\ge D\}$, form the far memory and are assigned the same virtual position $t-D$.
Their attention logits are
\begin{equation}
\tilde{s}_{p,j}
=
\frac{q_p^\top \mathbf{R}_{p-t+D} k_j}{\sqrt{d}},
\qquad
j\in\mathcal{F}_p .
\label{eq:far_logit}
\end{equation}
For $p=t+r$, where $r\in[0,B-1]$ is the block offset, every far entry is read at the effective distance $D+r\in[D,D+B-1]$, independent of input length.
FOLD thus removes exact long-range offsets without discarding any context entries.
For a fixed $p$, all entries in $\mathcal{F}_p$ share the same positional factor and are distinguished only by their keys $k_j$, whose underlying features $c_j$ can still encode coarse position (\S\ref{sec:diagnosis}).

The shared rotation depends only on the block offset $r$ and can be applied once to the query:
$q_p^\top\mathbf{R}_{D+r}k_j=\left(\mathbf{R}_{D+r}^\top q_p\right)^\top k_j $.
Far-memory keys can thus be cached unrotated and reused across rounds, while the position-dependent transformation is evaluated once per block query.
If the context contains at most $W$ entries, every $\mathcal{F}_p$ is empty and FOLD reduces exactly to the original drafter.

\paragraph{Decoupled memory normalization.}
FOLD computes attention over the window and far memory separately to control two effects of a growing memory.
First, attention within the memory can become diffuse as weakly relevant entries accumulate.
Second, the aggregate memory weight can grow with the number of far entries and overwhelm the recent window. 
The window branch retains the standard softmax over $\mathcal{N}_p$, with mass and output
\[
Z_{\mathcal{N}}
=
\sum_{j\in\mathcal{N}_p} e^{s_{p,j}},
\quad
o_{\mathcal{N}}
=
\sum_{j\in\mathcal{N}_p}
\frac{e^{s_{p,j}}}{Z_{\mathcal{N}}}v_j .
\]
For the far memory, FOLD introduces a concentration factor $\alpha\ge1$ that addresses both effects.
It sharpens the distribution within the memory,
\begin{equation}
o_{\mathcal{F}}
=
\sum_{j\in\mathcal{F}_p}
\frac{e^{\alpha\tilde{s}_{p,j}}}
     {\sum_{k\in\mathcal{F}_p}e^{\alpha\tilde{s}_{p,k}}}
v_j ,
\label{eq:sharpen}
\end{equation}
and defines its aggregate mass as the $\ell_\alpha$ norm of the exponentiated logits:
\begin{equation}
Z_{\mathcal{F}}
=
\Bigl(
\sum\nolimits_{j\in\mathcal{F}_p}
e^{\alpha\tilde{s}_{p,j}}
\Bigr)^{1/\alpha}.
\label{eq:far_mass}
\end{equation}
The two branches are combined by their masses:
\begin{equation}
o_p
=
\frac{
Z_{\mathcal{N}}o_{\mathcal{N}}
+
Z_{\mathcal{F}}o_{\mathcal{F}}
}{
Z_{\mathcal{N}}+Z_{\mathcal{F}}
}.
\label{eq:merge}
\end{equation}

When $\alpha=1$, FOLD reduces to a single softmax over its position-mapped logits.
For $\alpha>1$, it sharpens the far-memory read while tempering its aggregate mass, leaving the window logits unchanged.
This differs from ReRoPE's length-dependent scaling of all logits under a single softmax~\cite{su2023rerope}.
For $N=|\mathcal{F}_p|>0$ and maximum logit $\tilde{s}_{\max}$, Eq.~(\ref{eq:far_mass}) gives $e^{\tilde{s}_{\max}}\le Z_{\mathcal{F}}\le N^{1/\alpha}e^{\tilde{s}_{\max}}$.
The memory mass is at most $N^{1/\alpha}$ times that of its highest-scoring entry, compared with up to $N$ times under a single softmax.
Meanwhile, Eq.~(\ref{eq:sharpen}) concentrates the memory read on its highest-scoring entries.
As $\alpha\to\infty$, the memory approaches a hard top-1 read for a unique maximum and competes with the window as a single entry, without a learned scorer or discrete retrieval.

\paragraph{Properties.}
FOLD is training-free and leaves the target model and verification unchanged.
Every layer using FOLD reads its local window and the entire far memory, without retrieval or selection refresh.
For drafters mixing sliding-window and full-attention layers, FOLD can be applied to the full-attention layers alone or to all layers.
In the latter case, each sliding-window layer retains its native window as the local branch and additionally reads the far memory.

%% file: sections/04Experiment.tex
\section{Experiments}
\label{sec:experiments}

\subsection{Experimental Setup}
\label{sec:setup}

\paragraph{Models and drafters.}
We evaluate four target models with released block-parallel drafters of different designs (Table~\ref{tab:drafters}).
Llama-3.1-8B-Instruct~\cite{grattafiori2024llama3} and Qwen3.5-9B~\cite{qwen35} use the DFlash drafters of z-lab, and Gemma-4-12B-it~\cite{gemma4} and Qwen3-8B~\cite{yang2025qwen3} use the DFlash and DSpark~\cite{cheng2026dspark} drafters of DeepSpec~\cite{deepspec}.
For every drafter, FOLD uses the window $W$ in Table~\ref{tab:drafters}, fixed before evaluation, and $\alpha=2$.

\paragraph{Benchmarks.}
We evaluate seven benchmarks with inputs of about 1K to 256K tokens, grouped into general and code tasks.
The general tasks are GovReport from LongBench-E (Gov)~\cite{bai2024longbench}, LongBench~v2 (LBv2)~\cite{bai2025longbenchv2}, and five InfiniteBench~\cite{zhang2024infinitebench} tasks, namely KV retrieval (Retr.KV) and the English summarization (En.Sum), QA (En.QA), multiple choice (En.MC), and dialogue QA (En.Dia) tasks, whose average we report as En.
The code tasks are repository-level code continuation from Long Code Arena (LCA)~\cite{bogomolov2024long} at 32K, 64K, and 122K tokens, cross-file completion from RepoBench~\cite{liu2024repobench}, and patch generation from SWE-bench~\cite{jimenez2024swebench}.
For the length analysis, we add LCA inputs of 180K and 240K tokens and group LBv2 requests by length.
Requests exceeding a model's context limit are excluded (Appendix~\ref{app:data}).

\paragraph{Baselines.}
All baselines keep the original drafter and change only how it reads its context: StreamingLLM~\cite{xiao2024streamingllm} truncates the context, YaRN~\cite{peng2024yarn} and ReRoPE~\cite{su2023rerope} rescale or remap positions, and InfLLM~\cite{xiao2024infllm} and CMR, the cross-model retrieval mechanism of SpecExtend~\cite{cha2026specextend}, retrieve distant context.
We adapt ReRoPE and InfLLM from language models as baselines with closely related treatments of distant positions.
We call the original drafters DFlash and DSpark (Appendix~\ref{app:baselines}).

\paragraph{Metrics and protocol.}
We report acceptance length $\tau$ and the speedup over target-only AR decoding. Speedup is the mean per-request ratio $\mathrm{TPOT}_{\mathrm{AR}}/\mathrm{TPOT}_{\mathrm{method}}$, where the time per output token (TPOT) excludes prefill and each method is paired with an AR run on the same GPU.
Averages across benchmarks or tasks, such as Avg.\ and En., weight each equally.
We compute 95\% confidence intervals by bootstrapping over paired requests (Appendix~\ref{app:stats}).
All methods run in vLLM~\cite{kwon2023vllm} on one H100 GPU with batch size 1 and greedy decoding (Appendix~\ref{app:impl}).

\begin{table}[t]
\centering
\footnotesize
\setlength{\tabcolsep}{3pt}
\renewcommand{\arraystretch}{0.95}\setlength{\aboverulesep}{0.3ex}\setlength{\belowrulesep}{0.5ex}
\resizebox{\columnwidth}{!}{%
\begin{tabular}{@{}llcccc@{}}
\toprule
Target & Drafter & $B$ & Attention layers & Train & $W$ \\
\midrule
Llama-3.1-8B & DFlash & 10 & 5 full & 3K & 2,048 \\
Qwen3.5-9B & DFlash & 16 & 5 SW (4K) + 1 full & 40K & 4,096 \\
Gemma-4-12B & DFlash, DSpark & 7 & 5 full, partial RoPE & 4K & 4,096 \\
Qwen3-8B & DFlash, DSpark & 7 & 5 full & 4K & 2,048 \\
\bottomrule
\end{tabular}%
}
\caption{\textbf{Drafters.} Block size $B$, attention layers (SW: sliding window), training length, and the window $W$ of FOLD. Checkpoints are listed in Appendix~\ref{app:models}.}
\label{tab:drafters}
\end{table}

\begin{table*}[t]
\centering
\footnotesize
\setlength{\tabcolsep}{2.6pt}
\renewcommand{\arraystretch}{0.95}\setlength{\aboverulesep}{0.3ex}\setlength{\belowrulesep}{0.5ex}
\begin{tabular}{lcccccccccccccccc}
\toprule
\multirow{2}{*}{Method} & \multicolumn{2}{c}{Gov} & \multicolumn{2}{c}{LBv2} & \multicolumn{2}{c}{Retr.KV} & \multicolumn{2}{c}{En.} & \multicolumn{2}{c}{LCA} & \multicolumn{2}{c}{RepoBench} & \multicolumn{2}{c}{SWE-bench} & \multicolumn{2}{c}{Avg.} \\
\cmidrule(lr){2-3} \cmidrule(lr){4-5} \cmidrule(lr){6-7} \cmidrule(lr){8-9} \cmidrule(lr){10-11} \cmidrule(lr){12-13} \cmidrule(lr){14-15} \cmidrule(lr){16-17}
 & $\tau$ & Spd. & $\tau$ & Spd. & $\tau$ & Spd. & $\tau$ & Spd. & $\tau$ & Spd. & $\tau$ & Spd. & $\tau$ & Spd. & $\tau$ & Spd. \\
\midrule
\multicolumn{17}{l}{\textit{Llama-3.1-8B}} \\
DFlash & 2.49 & 2.35$\times$ & 2.88 & 2.62$\times$ & 1.35 & 1.24$\times$ & 1.77 & 1.64$\times$ & 2.75 & 2.53$\times$ & 2.87 & 2.68$\times$ & 1.86 & 1.74$\times$ & 2.28 & 2.11$\times$ \\
StreamingLLM & 2.88 & 2.71$\times$ & 3.88 & 3.57$\times$ & 1.79 & 1.69$\times$ & 2.70 & 2.57$\times$ & 4.06 & 3.79$\times$ & 4.07 & 3.79$\times$ & 2.36 & 2.22$\times$ & 3.11 & 2.91$\times$ \\
YaRN & 3.49 & 3.18$\times$ & 3.87 & 2.91$\times$ & 1.66 & 1.12$\times$ & 2.65 & 1.87$\times$ & 4.01 & 3.03$\times$ & 3.94 & 3.22$\times$ & 3.02 & 2.42$\times$ & 3.23 & 2.54$\times$ \\
CMR & 3.39 & 2.58$\times$ & 4.05 & 2.98$\times$ & 1.76 & 1.44$\times$ & 2.87 & 2.36$\times$ & 4.34 & 3.41$\times$ & 4.19 & 3.22$\times$ & 2.82 & 2.21$\times$ & 3.35 & 2.60$\times$ \\
ReRoPE & \textbf{3.64} & \textbf{3.35$\times$} & 4.14 & \textit{3.64$\times$} & \textit{1.90} & \textit{1.71$\times$} & \textit{3.20} & \textit{2.89$\times$} & 4.89 & \textit{4.40$\times$} & 4.40 & \textit{3.98$\times$} & \textit{3.62} & \textit{3.31$\times$} & \textit{3.69} & \textit{3.32$\times$} \\
InfLLM & 3.59 & 2.74$\times$ & \textbf{4.18} & 3.21$\times$ & \textit{1.90} & 1.59$\times$ & 3.13 & 2.58$\times$ & \textit{4.91} & 3.89$\times$ & \textit{4.41} & 3.41$\times$ & 3.50 & 2.75$\times$ & 3.66 & 2.88$\times$ \\
\textbf{FOLD (ours)} & \textit{3.62} & \textit{3.34$\times$} & \textit{4.17} & \textbf{3.67$\times$} & \textbf{2.21} & \textbf{1.99$\times$} & \textbf{3.23} & \textbf{2.92$\times$} & \textbf{5.03} & \textbf{4.52$\times$} & \textbf{4.46} & \textbf{4.04$\times$} & \textbf{3.65} & \textbf{3.33$\times$} & \textbf{3.77} & \textbf{3.40$\times$} \\
\midrule
\multicolumn{17}{l}{\textit{Qwen3.5-9B}} \\
DFlash & 3.22 & \textit{2.27$\times$} & 4.28 & 2.91$\times$ & 2.32 & 1.69$\times$ & 2.84 & 2.07$\times$ & 6.92 & 4.90$\times$ & 6.18 & 4.30$\times$ & 3.85 & 2.74$\times$ & 4.23 & 2.98$\times$ \\
StreamingLLM & 3.06 & 2.15$\times$ & 4.27 & 2.92$\times$ & 2.32 & 1.70$\times$ & 2.86 & 2.08$\times$ & 6.75 & 4.77$\times$ & 6.06 & 4.21$\times$ & 3.57 & 2.53$\times$ & 4.13 & 2.91$\times$ \\
YaRN & 3.22 & 2.26$\times$ & 4.36 & 2.91$\times$ & 2.40 & 1.69$\times$ & 2.98 & 2.15$\times$ & 7.19 & 5.07$\times$ & 6.24 & 4.34$\times$ & 3.93 & 2.79$\times$ & 4.33 & 3.03$\times$ \\
CMR & \textit{3.32} & 2.07$\times$ & 4.39 & 2.37$\times$ & \textbf{3.72} & \textit{2.30$\times$} & \textbf{3.27} & 2.12$\times$ & 7.23 & 4.52$\times$ & 6.21 & 3.81$\times$ & \textit{4.68} & 2.94$\times$ & \textit{4.69} & 2.88$\times$ \\
ReRoPE & \textbf{3.35} & \textbf{2.35$\times$} & 4.15 & 2.75$\times$ & 2.71 & 1.89$\times$ & 3.08 & \textit{2.21$\times$} & \textit{7.46} & \textit{5.21$\times$} & \textit{6.36} & 4.35$\times$ & 4.44 & \textit{3.14$\times$} & 4.51 & \textit{3.13$\times$} \\
InfLLM & 3.22 & 2.24$\times$ & \textit{4.42} & \textit{2.96$\times$} & 2.61 & 1.89$\times$ & 3.01 & 2.19$\times$ & 7.33 & 5.18$\times$ & \textit{6.36} & \textit{4.42$\times$} & 4.08 & 2.90$\times$ & 4.43 & 3.11$\times$ \\
\textbf{FOLD (ours)} & \textbf{3.35} & \textbf{2.35$\times$} & \textbf{4.62} & \textbf{3.03$\times$} & \textit{3.36} & \textbf{2.34$\times$} & \textit{3.23} & \textbf{2.32$\times$} & \textbf{8.06} & \textbf{5.67$\times$} & \textbf{6.70} & \textbf{4.64$\times$} & \textbf{4.75} & \textbf{3.35$\times$} & \textbf{4.87} & \textbf{3.39$\times$} \\
\midrule
\multicolumn{17}{l}{\textit{Gemma-4-12B}} \\
DFlash & \textit{3.04} & \textbf{2.79$\times$} & 2.82 & 2.49$\times$ & 2.57 & 2.24$\times$ & 2.00 & 1.77$\times$ & 2.67 & 2.41$\times$ & 2.74 & 2.49$\times$ & 3.31 & 3.00$\times$ & 2.73 & 2.46$\times$ \\
StreamingLLM & 2.72 & 2.50$\times$ & 2.69 & 2.41$\times$ & 2.12 & 1.83$\times$ & 2.18 & \textit{1.99$\times$} & 2.94 & \textbf{2.70$\times$} & \textit{3.13} & \textbf{2.86$\times$} & 2.63 & 2.42$\times$ & 2.63 & 2.39$\times$ \\
YaRN & \textbf{3.05} & \textit{2.76$\times$} & 2.91 & 2.34$\times$ & 2.89 & 2.05$\times$ & 2.15 & 1.63$\times$ & 2.73 & 2.26$\times$ & 2.78 & 2.39$\times$ & 3.38 & 2.85$\times$ & 2.84 & 2.33$\times$ \\
CMR & 2.91 & 2.36$\times$ & 2.96 & 2.35$\times$ & 2.47 & 1.92$\times$ & 2.39 & 1.94$\times$ & \textbf{3.14} & 2.56$\times$ & \textbf{3.18} & 2.59$\times$ & 3.26 & 2.67$\times$ & 2.90 & 2.34$\times$ \\
ReRoPE & \textbf{3.05} & \textit{2.76$\times$} & \textbf{3.12} & \textbf{2.71$\times$} & \textit{3.27} & \textit{2.69$\times$} & \textbf{2.42} & \textbf{2.09$\times$} & 3.03 & \textit{2.67$\times$} & 3.00 & 2.67$\times$ & \textit{3.50} & \textit{3.11$\times$} & \textit{3.05} & \textit{2.67$\times$} \\
InfLLM & \textit{3.04} & 2.38$\times$ & 3.01 & 2.32$\times$ & 2.83 & 2.14$\times$ & 2.27 & 1.79$\times$ & 3.02 & 2.38$\times$ & 3.00 & 2.35$\times$ & 3.45 & 2.72$\times$ & 2.94 & 2.30$\times$ \\
\textbf{FOLD (ours)} & 3.03 & 2.74$\times$ & \textit{3.11} & \textit{2.68$\times$} & \textbf{3.32} & \textbf{2.71$\times$} & \textit{2.41} & \textbf{2.09$\times$} & \textit{3.06} & \textbf{2.70$\times$} & 3.04 & \textit{2.70$\times$} & \textbf{3.51} & \textbf{3.13$\times$} & \textbf{3.07} & \textbf{2.68$\times$} \\
\bottomrule
\end{tabular}
\caption{\textbf{Main results with DFlash drafters.} Acceptance length $\tau$ and speedup (Spd.) over AR decoding. Layer coverage for Qwen3.5-9B is specified in Appendix~\ref{app:baselines}. Bold and italics mark the best and second-best values per column and model.}
\label{tab:main}
\end{table*}

\subsection{Main Results}
\label{sec:main_results}

\paragraph{Overall performance.}
FOLD achieves the highest average $\tau$ and speedup on all three models (Table~\ref{tab:main}), raising the average speedup of the original drafters from 2.11$\times$ to 3.40$\times$ on Llama-3.1-8B, from 2.98$\times$ to 3.39$\times$ on Qwen3.5-9B, and from 2.46$\times$ to 2.68$\times$ on Gemma-4-12B.
The gain is largest on Llama-3.1-8B, whose drafter degrades most on long inputs.
FOLD and ReRoPE, which both retain the full context while bounding its positions, rank first and second in average speedup on all three models.
The retrieval baselines can attain comparable or higher $\tau$, as CMR does on Retr.KV for Qwen3.5-9B, but their selection and refresh overhead lowers their speedup.
On Retr.KV, where drafts must reproduce values stored in distant context, FOLD is the fastest method on all three models.
On the Gov, LBv2, Retr.KV, and En.Sum requests of Llama-3.1-8B, an SD round takes 12.3\,ms with FOLD, 0.3\,ms more than with DFlash and less than with InfLLM (13.7\,ms) and CMR (14.0\,ms).
The side cache is at most about 6.0\,GiB, on Qwen3.5-9B at its maximum context length with all six drafter layers reading the memory (Appendix~\ref{app:impl}).

\begin{table}[t]
\centering
\footnotesize
\setlength{\tabcolsep}{2.7pt}
\renewcommand{\arraystretch}{0.95}\setlength{\aboverulesep}{0.3ex}\setlength{\belowrulesep}{0.5ex}
\begin{tabular*}{\columnwidth}{@{\extracolsep{\fill}}lcccccc@{}}
\toprule
\multirow{3}{*}{Method} & \multicolumn{2}{c}{Qwen3-8B} & \multicolumn{4}{c}{Gemma-4-12B} \\
\cmidrule(lr){2-3}\cmidrule(lr){4-7}
 & \multicolumn{2}{c}{General} & \multicolumn{2}{c}{General} & \multicolumn{2}{c}{Code} \\
\cmidrule(lr){2-3}\cmidrule(lr){4-5}\cmidrule(lr){6-7}
 & $\tau$ & Spd. & $\tau$ & Spd. & $\tau$ & Spd. \\
\midrule
DSpark & 4.02 & 3.41$\times$ & 3.03 & 2.62$\times$ & 3.68 & 3.22$\times$ \\
StreamingLLM & 4.19 & 3.54$\times$ & 2.68 & 2.33$\times$ & 3.54 & 3.13$\times$ \\
YaRN & 4.27 & 3.32$\times$ & 3.21 & 2.45$\times$ & 3.72 & 3.03$\times$ \\
CMR & 4.25 & 2.91$\times$ & 2.94 & 2.25$\times$ & 3.79 & 3.00$\times$ \\
ReRoPE & 4.32 & 3.59$\times$ & 3.38 & 2.79$\times$ & \textbf{3.95} & \textbf{3.38$\times$} \\
InfLLM & \textbf{4.39} & 3.11$\times$ & 3.14 & 2.36$\times$ & 3.91 & 2.98$\times$ \\
\textbf{FOLD (ours)} & 4.38 & \textbf{3.63$\times$} & \textbf{3.42} & \textbf{2.83$\times$} & \textbf{3.95} & \textbf{3.38$\times$} \\
\bottomrule
\end{tabular*}
\caption{\textbf{Results with DSpark drafters.}
Acceptance length $\tau$ and speedup over AR decoding on the general and code tasks. General includes only inputs of at least 16K tokens and, for Qwen3-8B, only Gov and LBv2.}
\label{tab:extra}
\end{table}

\begin{figure*}[t]
  \centering
  \includegraphics[width=\textwidth]{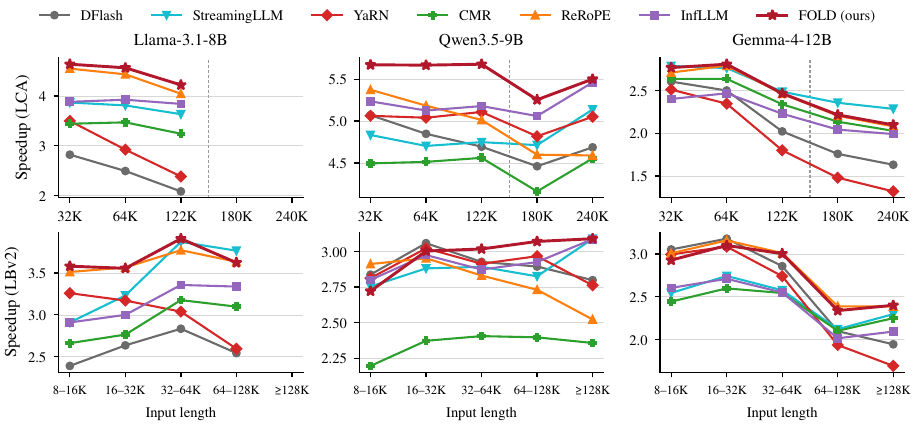}
  \caption{\textbf{Speedup over AR by input length.} Top: LCA inputs of fixed length, where the 180K and 240K inputs come from its huge-context split (dotted line). Bottom: LBv2 requests grouped by length. Llama-3.1-8B does not admit inputs beyond 128K. Appendix~\ref{app:length_tau} shows the corresponding $\tau$.}
  \label{fig:length}
\end{figure*}

\paragraph{Effect of input length.}
\label{sec:length}
Figure~\ref{fig:length} breaks the speedup down by input length on LCA (32K--240K tokens) and LBv2 (10K--250K tokens).
On LCA, the original drafters slow down as the input grows, and the gain of FOLD over them widens from 32K to 122K tokens on all three models.
On Llama-3.1-8B, FOLD raises the speedup from 2.82$\times$ to 4.64$\times$ at 32K tokens and from 2.08$\times$ to 4.22$\times$ at 122K, and on Qwen3.5-9B it stays at 5.67--5.68$\times$ from 32K to 122K.
Among the baselines, YaRN declines fastest on Llama-3.1-8B and Gemma-4-12B, and ReRoPE declines with length on Qwen3.5-9B, from 5.37$\times$ to 4.59$\times$ on LCA and to 2.52$\times$ beyond 128K tokens on LBv2.
The relative performance of StreamingLLM, which keeps only recent entries, depends strongly on the model: it is the fastest method at four of the five LCA lengths on Gemma-4-12B but slower than DFlash on Qwen3.5-9B up to 64K tokens.
Where DFlash barely degrades, as on LBv2 inputs below 16K tokens for Qwen3.5-9B and Gemma-4-12B, FOLD has little to gain and is slightly slower.
We also report results on SPEED-Bench (Appendix~\ref{app:length_tau}).

\paragraph{Other drafters and engines.}
With DSpark drafters (Table~\ref{tab:extra}), FOLD achieves the highest speedup on the general tasks of both models, raising it from 3.41$\times$ to 3.63$\times$ on Qwen3-8B and from 2.62$\times$ to 2.83$\times$ on Gemma-4-12B.
On the code tasks of Gemma-4-12B, FOLD and ReRoPE share the highest speedup of 3.38$\times$, compared with 3.22$\times$ for DSpark.
Results for Qwen3-8B with DFlash, evaluated up to 32K tokens, are reported in Appendix~\ref{app:qwen3_dflash}.
The gains also extend to SGLang, where FOLD raises the average speedup on the general tasks of Llama-3.1-8B from 1.54$\times$ to 2.50$\times$, the highest among all methods (Appendix~\ref{app:sglang}).

\subsection{Analysis}
\label{sec:analysis}

\begin{table}[t]
\centering
\footnotesize
\setlength{\tabcolsep}{3.5pt}
\renewcommand{\arraystretch}{0.95}\setlength{\aboverulesep}{0.3ex}\setlength{\belowrulesep}{0.5ex}
\begin{tabular}{@{}lcccccc@{}}
\toprule
 & \multicolumn{3}{c}{DFlash $\rightarrow$ Window} & \multicolumn{3}{c}{Window $\rightarrow$ FOLD} \\
\cmidrule(lr){2-4}\cmidrule(l){5-7}
 & 32K & 64K & 122K & 32K & 64K & 122K \\
\midrule
Llama & $+1.07$ & $+1.31$ & $+1.53$ & $+1.02$ & $+1.07$ & $+0.93$ \\
Qwen3.5 & $-0.22$ & $-0.08$ & $+0.15$ & $+1.11$ & $+1.28$ & $+1.29$ \\
Gemma & $+0.09$ & $+0.13$ & $+0.36$ & $+0.15$ & $+0.29$ & $+0.21$ \\
\bottomrule
\end{tabular}
\caption{\textbf{Decomposition of the difference in $\tau$} between FOLD and DFlash on LCA into Window minus DFlash and FOLD minus Window. For Qwen3.5-9B, Window restricts only the full-attention layer, while FOLD lets all six layers read distant entries.}
\label{tab:decomp}
\end{table}


\paragraph{Window and memory.}
To examine the roles of the window and the memory, we compare FOLD with Window, which keeps the window of FOLD but drops the distant entries (Table~\ref{tab:decomp}).
On LCA, FOLD exceeds Window by 0.15--1.29 in $\tau$ for all three drafters and all three lengths.
The window alone behaves differently across drafters. It raises $\tau$ by 1.07--1.53 on Llama-3.1-8B, more on longer inputs, and by 0.09--0.36 on Gemma-4-12B, whereas on Qwen3.5-9B it lowers $\tau$ by 0.22 and 0.08 at 32K and 64K and raises it by 0.15 at 122K.
The memory contributes 0.93--1.07 on Llama-3.1-8B. On Qwen3.5-9B it contributes 1.11--1.29 and thus most or all of the gain over DFlash, suggesting that distant context can aid drafting without retaining its exact relative positions.
On LBv2, the memory raises $\tau$ on Gemma-4-12B and Qwen3.5-9B, whereas its gain shrinks with length on Llama-3.1-8B and it lowers $\tau$ on Qwen3-8B (Appendix~\ref{app:decomp}).

\begin{table}[t]
\centering
\footnotesize
\setlength{\tabcolsep}{3pt}
\renewcommand{\arraystretch}{0.95}\setlength{\aboverulesep}{0.3ex}\setlength{\belowrulesep}{0.5ex}
\resizebox{\columnwidth}{!}{%
\begin{tabular}{@{}llccc@{}}
\toprule
Model & Tasks & $\tau_{\alpha=2}$ & $\alpha=2$ vs.\ 1 & $\alpha=4$ vs.\ 2 \\
\midrule
Llama & General & 3.17 & 1.150 {\scriptsize[1.137, 1.162]} & 1.018$^\dagger$ {\scriptsize[1.014, 1.022]} \\
 & Code & 4.52 & 1.067 {\scriptsize[1.064, 1.072]} & 0.992 {\scriptsize[0.991, 0.994]} \\
\addlinespace[0.35em]
Qwen3 & General & 3.26 & 1.160 {\scriptsize[1.138, 1.183]} & 1.012$^\dagger$ {\scriptsize[1.006, 1.019]} \\
\addlinespace[0.35em]
Qwen3.5 & General & 3.59 & 2.793 {\scriptsize[2.743, 2.845]} & 1.146 {\scriptsize[1.133, 1.159]} \\
 & Code & 6.86 & 4.173 {\scriptsize[4.068, 4.285]} & 1.001 {\scriptsize[1.000, 1.003]} \\
\addlinespace[0.35em]
Gemma & General & 2.88 & 1.071 {\scriptsize[1.066, 1.077]} & 1.001$^\dagger$ {\scriptsize[0.998, 1.004]} \\
 & Code & 3.13 & 1.028 {\scriptsize[1.024, 1.032]} & 1.000 {\scriptsize[0.997, 1.002]} \\
\bottomrule
\end{tabular}%
}
\caption{\textbf{Effect of the concentration factor $\alpha$} on requests longer than the window. General and Code follow the task groups of \S\ref{sec:setup}, and for Qwen3-8B, General covers only Gov and LBv2. Both columns are means of per-request ratios over all requests of a group, with 95\% bootstrap intervals, and values above 1 favor the larger $\alpha$. $^\dagger$The ratio of the overall $\tau$ of all requests leads to a different conclusion (Appendix~\ref{app:alpha}).}
\label{tab:alpha}
\end{table}

\paragraph{Weighting the memory.}
Setting $\alpha=1$ keeps the position mapping of FOLD but normalizes the window and the memory in a single softmax.
For $N$ equally scored distant entries, the memory mass then scales linearly with $N$, compared with $\sqrt{N}$ under the default $\alpha=2$ (\S\ref{sec:fold}).
Increasing $\alpha$ both sharpens attention within the memory and tempers its aggregate mass relative to the window.
Raising $\alpha$ from 1 to 2 yields mean per-request $\tau$ ratios above 1 across all task groups in Table~\ref{tab:alpha}, with the largest gains on Qwen3.5-9B, where the ratios reach 2.79 on the general tasks and 4.17 on the code tasks.
On this drafter, both settings read the same distant entries in all six layers.
Yet on LBv2, $\alpha=1$ yields a $\tau$ of only 1.65, below DFlash's 4.28, whereas $\alpha=2$ reaches 4.62.
These results indicate strong sensitivity to memory normalization.
Since ReRoPE also normalizes the window and distant entries jointly, this sensitivity may help explain its decline with input length on this drafter (Figure~\ref{fig:length}).
Raising $\alpha$ further from 2 to 4 keeps the mean ratio within 2\% of 1 in no consistent direction on the other task groups, but yields 1.146 on Qwen3.5-9B's general tasks.
A larger $\alpha$ can therefore provide further gains, although its benefit depends on both the drafter and the task.

\paragraph{Gains across drafters.}
FOLD complements long-context training.
Although the Qwen3.5-9B drafter is trained on 40K-token sequences with sliding windows, FOLD raises its $\tau$ on all seven benchmarks.
Training length alone does not explain the variation in gains.
The Llama-3.1-8B and Gemma-4-12B drafters are both trained on short sequences, yet FOLD raises their $\tau$ by 0.86--2.28 and by at most 0.75.
Beyond distances and distant information, which FOLD addresses, drafting is also affected by the target itself, whose features and predictions change with input length.
On LCA, even StreamingLLM, whose positions are bounded, loses $\tau$ as inputs grow, suggesting that positional extrapolation is not the only cause of degradation (Appendix~\ref{app:length_tau}).
Differences in target model, training data, and drafter architecture may thus all contribute.
Across these differences, FOLD attains the highest average speedup on all three models in Table~\ref{tab:main} and the highest or tied-highest speedup with DSpark drafters (Table~\ref{tab:extra}).

%% file: sections/05Conclusion.tex
\section{Conclusion}
\label{sec:conclusion}

Our results show that drafters trained on short sequences can accelerate long-context decoding without retraining.
Because a drafter reads target hidden states computed over the full prefix, FOLD keeps exact positions only within a local window and reads older entries as a far memory whose weight a concentration factor controls.
On diverse drafters and inputs of up to 256K tokens, it consistently improves SD and is the most robust method across models.
Drafters can thus inherit the long-context ability of their target models, which shifts the open question from how distant entries are positioned to how they are attended to.
FOLD controls how much weight they receive, and richer ways of reading the memory may exploit it further.

%% file: sections/Appendix.tex
\appendix

\section{Details of the Diagnostic Experiments}
\label{app:fixed_state}

\paragraph{Offline acceptance length.}
For each input, we take the greedy continuation of $\mathcal{T}$, generated beforehand with vLLM, and obtain the context features $c_j$ at all positions from a single forward pass of $\mathcal{T}$ over the input and the continuation.
At a position $t$ of the continuation, each variant drafts a block from the features of $x_{<t}$, so all variants are compared on identical states.
Acceptance length $\tau_{\mathrm{fs}}$ is one plus the number of leading draft tokens that match the continuation, averaged over positions and then over inputs.
All diagnostics use a PyTorch~\cite{paszke2019pytorch} reference implementation of the drafter without any component of FOLD.

\paragraph{Distance control (Finding~1).}
We evaluate 16 SPEED-Bench 2K inputs at two decoding positions per input.
At each position, we shift the position IDs of the entire draft block forward by 8K, 32K, or 128K while leaving those of the committed entries unchanged.
This intervention increases the distance from the draft block to every context entry while preserving both the content and all within-block distances.
We report the draft acceptance fraction $a=\min(L,K)/K$, where $L$ is the number of leading matches and $K$ is the number of scored draft tokens: 9 for Llama-3.1-8B and 7 for Gemma-4-12B.
Unlike $\tau_{\mathrm{fs}}$, $a$ excludes the additional token sampled by $\mathcal{T}$.
For Gemma-4-12B, the paired-bootstrap 95\% confidence intervals for the change in $a$ relative to the unshifted baseline lie entirely below zero at all three shifts.

\paragraph{Position probe (Finding~2).}
For each of SPEED-Bench 32K and LongBench~v2, we evaluate 20 Llama-3.1-8B inputs using five-fold cross-validation over inputs, with 16 inputs for training and four for testing in each fold.
Outside the most recent 2,048 entries, we sample 1,500 pairs per input from each of five separation bins spanning 64 to 64K tokens.
One LongBench~v2 input contains no eligible pair in the farthest bin and is omitted from that bin only.
A logistic-regression probe, trained with Adam for 600 steps at a learning rate of $10^{-3}$, takes the standardized difference $c_i-c_j$, or the corresponding difference of token embeddings, and predicts whether $i<j$.
We report test accuracy averaged across folds, with a standard deviation of at most 1.8 percentage points for separations of at least 16K tokens.
For the shuffled control, we permute the older context in 64-token chunks before it is processed by $\mathcal{T}$, then realign each resulting feature with its original token.
The probe recovers the processing order with 84.0\% accuracy on SPEED-Bench 32K and 85.6\% on LongBench~v2, but recovers the original order with only 52.1\% and 52.2\%, respectively.
Original-order accuracy rises to 56--58\% in the farthest bin, possibly because of topical drift within documents.

\section{Experimental Details}
\label{app:exp}

\subsection{Models and Drafters}
\label{app:models}

Table~\ref{tab:checkpoints} lists the released checkpoints.
Every drafter is used as released, with its own block size and target-layer selection, and no drafter is fine-tuned.
The DeepSpec drafters of Gemma-4-12B and Qwen3-8B were trained with a maximum sequence length of 4,096 tokens~\cite{deepspec}, the z-lab drafter of Qwen3.5-9B on 40K-token sequences with sliding-window attention, and the z-lab drafter of Llama-3.1-8B with a maximum sequence length of 3,072 tokens~\cite{chen2026dflash}.

\paragraph{Licenses.}
All datasets and models are publicly released, and we use them only for research evaluation.
LongBench, InfiniteBench, and SWE-bench are released under MIT, LongBench~v2, Long Code Arena, Qwen3.5-9B, Qwen3-8B, and Gemma-4-12B under Apache-2.0, RepoBench under CC BY 4.0, SPEED-Bench under the NVIDIA Evaluation Dataset License, and Llama-3.1-8B-Instruct under the Llama 3.1 Community License.
The z-lab drafters are released under MIT and Apache-2.0, and the DeepSpec drafters accompany the MIT-licensed DeepSpec code.

\begin{table}[h]
\centering
\scriptsize
\setlength{\tabcolsep}{3pt}
\resizebox{\columnwidth}{!}{%
\begin{tabular}{ll}
\toprule
Target & Drafter checkpoint \\
\midrule
NousResearch/Meta-Llama-3.1-8B-Instruct & z-lab/LLaMA3.1-8B-Instruct-DFlash-UltraChat \\
Qwen/Qwen3.5-9B & z-lab/Qwen3.5-9B-DFlash \\
google/gemma-4-12B-it & deepseek-ai/dflash\_gemma4\_12b\_block7 \\
google/gemma-4-12B-it & deepseek-ai/dspark\_gemma4\_12b\_block7 \\
Qwen/Qwen3-8B & deepseek-ai/dflash\_qwen3\_8b\_block7 \\
Qwen/Qwen3-8B & deepseek-ai/dspark\_qwen3\_8b\_block7 \\
\bottomrule
\end{tabular}%
}
\caption{Released checkpoints on Hugging Face.}
\label{tab:checkpoints}
\end{table}

\subsection{Evaluation Data}
\label{app:data}

Table~\ref{tab:data} summarizes the evaluation data.
For each model, all methods are evaluated on the same requests with identical input token sequences, and generation ends at the target's end-of-turn token or at the generation limit.

\paragraph{General tasks.}
Prompts are formatted as a single user turn with the target's chat template, with thinking disabled for Qwen models.
Gov uses the official LongBench prompt, LBv2 the official zero-shot prompt of LongBench~v2, and Retr.KV and En.Sum the official InfiniteBench prompts.
For En.QA, En.MC, and En.Dia, we replace the answer-format instruction of the official InfiniteBench prompts with an instruction to reason step by step before stating the final answer, and generate up to 512 tokens in a single call.

\paragraph{Code tasks.}
LCA uses the large-context split of the Long Code Arena project-level code completion benchmark~\cite{bogomolov2024long}, which pairs a Python file with a snapshot of its repository, and its huge-context split for the 180K and 240K inputs.
Each file is cut before the first line that contains a function or class from the repository and is preceded by at least 50 lines.
The input places the other Python files of the repository before the prefix of the cut file, each headed by a comment with its path, with files nearer in the directory tree placed closer to the prefix.
Files are added from the nearest until the input reaches exactly 32K, 64K, 122K, 180K, or 240K tokens, and the farthest added file is truncated from its start.
We keep every eligible example of the large-context split and the first 48 eligible examples per length of the huge-context split.
RepoBench uses the cross-file-first split of RepoBench~v1.1~\cite{liu2024repobench} for Python and Java, in which the line to complete is the first use of a definition from another file, with the official prompt that places code snippets from other files before the imports and preceding lines of the current file.
We take its 32k, 64k, and 128k levels, which the benchmark defines with its own tokenizer, keep one example per file, and use all Python files and a random sample of 134 Java files per level.
SWE-bench uses a random sample of 288 test instances of SWE-bench~\cite{jimenez2024swebench} with the BM25-retrieved code context of its 40K-token setting, from which the target writes a patch that resolves the issue.
LCA and RepoBench inputs are raw text without a chat template, whereas SWE-bench prompts use the chat template of the general tasks.
Random samples take the first examples in a fixed hash order of their identifiers.

\paragraph{SPEED-Bench.}
SPEED-Bench~\cite{abramovich2026speedbench} is not one of the seven benchmarks of Section~\ref{sec:setup}.
We use its throughput tiers of 1K to 32K tokens, with the first turn of each item as the prompt, for the diagnostics of Section~\ref{sec:diagnosis}, Figure~\ref{fig:length_speedup_down}, the length dispatch check in Appendix~\ref{app:impl}, and the supplementary results in Tables~\ref{tab:speed} and~\ref{tab:alpha_full}.

\paragraph{Exclusions and subsets.}
We compute the required context length by summing the input length, generation limit, and a 65-token margin, then rounding the total up to a multiple of 256.
Requests whose required context length exceeds the model's maximum context length are excluded, resulting in slightly different sample counts across models.
Requests of the general tasks that were used during method development or pipeline checks are also excluded.
One Gemma-4-12B RepoBench request, for which the target stops after its first output token, has no TPOT and is excluded for all methods.
Qwen3-8B with DFlash is evaluated on 283 Gov and 79 LBv2 requests.
With the DSpark drafters, the general tasks keep only inputs of at least 16K tokens, which leaves 31 Gov and 63 LBv2 requests for Qwen3-8B and 32 Gov, 356 LBv2, 478 Retr.KV, 72 En.Sum, 281 En.QA, 181 En.MC, and 200 En.Dia requests for Gemma-4-12B.
On the code tasks, the DSpark drafter of Gemma-4-12B uses the same requests as its DFlash drafter, and Qwen3-8B is not evaluated because most of these inputs exceed its 32K context.

\begin{table*}[t]
\centering
\small
\begin{tabular}{lccccccc}
\toprule
 & & \multicolumn{2}{c}{Llama-3.1-8B} & \multicolumn{2}{c}{Qwen3.5-9B} & \multicolumn{2}{c}{Gemma-4-12B} \\
\cmidrule(lr){3-4}\cmidrule(lr){5-6}\cmidrule(lr){7-8}
Task & Gen.\ limit & $n$ & Input length & $n$ & Input length & $n$ & Input length \\
\midrule
Gov & 512 & 284 & 1.8K--27K & 275 & 1.1K--28K & 275 & 1.1K--28K \\
LBv2 & 128 & 320 & 9.8K--128K & 379 & 10K--253K & 378 & 10K--250K \\
Retr.KV & 128 & 488 & 121K--122K & 478 & 165K--166K & 478 & 170K--170K \\
En.Sum & 1200 & 48 & 65K--125K & 72 & 66K--235K & 72 & 66K--240K \\
En.QA & 512 & 112 & 64K--128K$^\dagger$ & 281 & 64K--256K$^\dagger$ & 281 & 64K--256K$^\dagger$ \\
En.MC & 512 & 79 & 64K--128K$^\dagger$ & 181 & 64K--256K$^\dagger$ & 181 & 64K--256K$^\dagger$ \\
En.Dia & 512 & 200 & 64K--128K$^\dagger$ & 200 & 64K--128K$^\dagger$ & 200 & 64K--128K$^\dagger$ \\
\midrule
LCA 32K & 256 & 236 & 32K & 236 & 32K & 236 & 32K \\
LCA 64K & 256 & 222 & 64K & 224 & 64K & 232 & 64K \\
LCA 122K & 256 & 129 & 122K & 154 & 122K & 176 & 122K \\
LCA 180K & 256 & -- & -- & 48 & 180K & 48 & 180K \\
LCA 240K & 256 & -- & -- & 48 & 240K & 48 & 240K \\
RepoBench Python & 256 & 393 & 8.2K--85K & 393 & 8.7K--92K & 393 & 6.8K--106K \\
RepoBench Java & 256 & 402 & 7.5K--86K & 402 & 8.5K--110K & 401 & 9.3K--122K \\
SWE-bench & 512 & 288 & 31K--39K & 288 & 38K--51K & 288 & 41K--56K \\
\midrule
SPEED 1K & 1024 & 256 & -- & 256 & -- & 256 & -- \\
SPEED 2K & 1024 & 256 & -- & 256 & -- & 256 & -- \\
SPEED 8K & 1024 & 256 & -- & 256 & -- & 256 & -- \\
SPEED 16K & 1024 & 768 & -- & 754 & -- & 751 & -- \\
SPEED 32K & 1024 & 768 & -- & 761 & -- & 768 & -- \\
\bottomrule
\end{tabular}
\caption{\textbf{Evaluation data.} Number of requests $n$, input length range in tokens, and generation limit per task. $^\dagger$Range of the length bins that contain the requests. LCA inputs have exactly the stated length, and SPEED-Bench inputs have the nominal length of their tier.}
\label{tab:data}
\end{table*}

\subsection{Baseline Configurations}
\label{app:baselines}

All baselines modify only how the drafter attends to its context and leave the target model and verification unchanged.
\begin{itemize}
  \item \textbf{StreamingLLM}~\cite{xiao2024streamingllm} keeps the first 4 context entries as attention sinks and the 2,044 most recent entries (2,048 in total). The retained entries are renumbered consecutively, and the block positions follow them.
  \item \textbf{YaRN}~\cite{peng2024yarn} attends to the full context with a per-request scaling factor $\max\big(1, \lceil (P+G+Q)/256\rceil \cdot 256 / R\big)$, where $P$ is the input length, $G$ the generation limit, $Q$ the number of block positions, and $R$ the reference length (3,072 for Llama-3.1-8B, 40,000 for Qwen3.5-9B, and 4,096 for Gemma-4-12B and Qwen3-8B).
  \item \textbf{ReRoPE}~\cite{su2023rerope} keeps true distances within the window $\mathcal{N}_p$ of FOLD, assigns every older entry the distance $D$ at all block positions, and scales every query in every drafter layer by $\max\big(1, \log(p+1)/\log 4096\big)$. The window and the older entries are normalized by a single softmax.
  \item \textbf{CMR} is the cross-model retrieval of SpecExtend~\cite{cha2026specextend}, adapted to block-parallel drafting with the defaults of its official EAGLE implementation. The context is divided into consecutive chunks of 32 entries. The first draft block after prefill reads the 64 most recent chunks, and the selection is refreshed every 8 rounds. At a refresh for anchor $x_t$, CMR averages over heads the attention weights of the target query at position $t-1$ in the target's last full-attention layer, taken from the preceding verification, and selects the 64 chunks with the highest mean weight (2,048 entries). Between refreshes, each newly started chunk is added to the selection and extended as entries are committed. The selected entries are renumbered consecutively in chronological order, and the $B$ block positions follow them. All drafter layers read the same selection, including the sliding-window layers of the Qwen3.5-9B drafter.
  \item \textbf{InfLLM}~\cite{xiao2024infllm} follows its official implementation and configuration (blocks of 128 entries, 128 initial entries, 16 retrieved blocks, and 4 representative entries per block) and uses the window $\mathcal{N}_p$ of FOLD as its local window. Entries that leave the window are grouped into blocks. Each block is represented by the mean key of the 4 entries that received the most attention while inside the window, and blocks whose entries never entered the window, which include most of a long input, use the mean key of all their entries. In every round and layer, the block queries, averaged over positions, select the 16 blocks whose representations have the largest inner product with them. The initial entries, the selected blocks, and the older entries not yet grouped into a block are all placed at the largest local distance $D-1$ and attended together with the window under a single softmax. Memory is kept on the GPU without CPU offloading.
\end{itemize}
StreamingLLM and CMR use a budget of 2,048 entries for all models, as in their standard configurations, whereas FOLD, ReRoPE, and InfLLM use the window $W$ of each model.
The ablation variant Window used in \S\ref{sec:analysis} attends only to the window $\mathcal{N}_p$ of FOLD, i.e., FOLD without its far memory.
The Qwen3.5-9B drafter reads the full context only in its last layer, while its five sliding-window layers read the 4,096 most recent entries.
FOLD and ReRoPE are applied to all six layers, so that each sliding-window layer keeps its native sliding window for nearby entries and additionally reads the older entries at the shared distant position.
StreamingLLM, YaRN, and InfLLM are applied to the full-attention layer alone, since the sliding-window layers never exceed their trained distances and InfLLM would otherwise retrieve in every layer.
These choices also gave the higher speedup in our evaluation.
CMR retrieves on the target side and, as in its original design, every drafter layer reads only the selected entries.
For this drafter, DFlash and YaRN use the storage-matched implementation described in Appendix~\ref{app:impl}.
The DSpark drafters use the same configurations as the DFlash drafters of the same target.

\subsection{Implementation}
\label{app:impl}

All methods run in vLLM~0.28.0 on a single NVIDIA H100 80GB HBM3 GPU with batch size 1, BF16 precision, CUDA graphs, chunked prefill of 8,192 tokens, and greedy decoding.
Each method is timed against an AR run of the same requests on the same GPU.
For each attention read of the drafter, the number of FlashAttention~\cite{dao2022flashattention} splits is chosen from $\{1, 8, 16, 32, 64\}$ by a microbenchmark on synthetic inputs with the actual attention shapes of that read. Methods that use the same read, namely FOLD, Window, ReRoPE, and the local read of InfLLM, share the chosen value.

\paragraph{FOLD.}
FOLD reads the window with one FlashAttention call over the drafter's KV cache and the far memory with a second call over a side cache, which stores the unrotated keys and the values of all committed entries.
The side cache is filled as entries are committed, and its keys are never re-rotated.
Fused Triton~\cite{tillet2019triton} kernels rotate the queries and merge the two outputs (Eqs.~\ref{eq:sharpen}--\ref{eq:merge}), and all kernels run within CUDA graphs.
At the maximum context length, the side cache takes 2.5\,GiB in BF16 for Llama-3.1-8B and Gemma-4-12B, 0.63\,GiB for Qwen3-8B, and 6.0\,GiB for Qwen3.5-9B, or 1.0\,GiB when only its full-attention layer reads the memory, which lowers its speedup averaged over the seven general tasks from 2.43$\times$ to 2.33$\times$.

\paragraph{Length dispatch.}
With length dispatch, the drafter runs a separate window-only CUDA graph while the sequence is no longer than $W$ and switches to the FOLD graph once the sequence grows beyond $W$.
Because the side cache is updated in every round, the far memory is complete at the switch.
We evaluate the dispatch on the 1K and 2K tiers and the first turns of the qualitative split of SPEED-Bench.
Requests that stay within the window produce the same outputs and $\tau$ as the original drafter and run about 1\% slower on Llama-3.1-8B, Qwen3-8B, and Gemma-4-12B.

\paragraph{Storage-matched baselines for Qwen3.5-9B.}
For the Qwen3.5-9B drafter, the stock DFlash implementation in vLLM uses different KV-cache storage, grouping, and attention wrappers from FOLD.
To control for these implementation differences, we evaluate DFlash and YaRN for this model using the same KV-cache storage and grouping scheme and the same FlashAttention wrapper as FOLD.

\subsection{Statistics}
\label{app:stats}

For each request, acceptance length $\tau$ is the average number of tokens committed per verification round, including the token sampled by $\mathcal{T}$.
We report its mean across requests.
Time per output token (TPOT) is the decoding time after the first output token divided by the number of subsequent output tokens, excluding prefill.
We compute speedup for each request as $\mathrm{TPOT}_{\mathrm{AR}}/\mathrm{TPOT}_{\mathrm{method}}$ and report the mean of these ratios across requests.
Results are averaged with equal weight across benchmarks, and En.\ averages its four tasks with equal weight.
The side-cache comparison in Appendix~\ref{app:impl} and the SGLang results in Appendix~\ref{app:sglang} instead weight the seven general tasks equally.
For paired comparisons, we report 95\% bootstrap confidence intervals using 2,000 resamples of paired requests with replacement and a fixed random seed.
For averages across tasks, requests are resampled separately within each task.

\paragraph{Output agreement.}
Greedy verification preserves the AR output in exact arithmetic, but in BF16 the verification pass over a block and the AR step can differ numerically, which can change the output where the two most likely tokens nearly tie.
Such differences affect every method, including the original drafters, so the reported speedups compare decoding on the same inputs rather than on identical outputs.

\begin{figure*}[t]
  \centering
  \includegraphics[width=\textwidth]{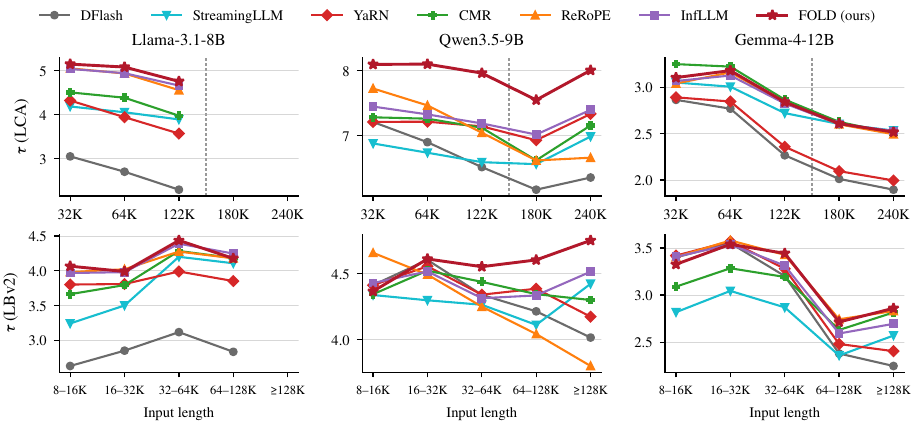}
  \caption{\textbf{Acceptance length by input length}, on the same requests and groups as Figure~\ref{fig:length}. Top: LCA. Bottom: LBv2.}
  \label{fig:length_tau}
\end{figure*}

\section{Additional Results}
\label{app:results}

\subsection{Results by Input Length}
\label{app:length_tau}

Figure~\ref{fig:length_tau} shows the acceptance length on the same requests and groups as Figure~\ref{fig:length}.
On LCA, from 32K to 122K tokens, the $\tau$ of DFlash falls by 10--25\% across the three drafters, that of StreamingLLM by 4--11\%, and that of FOLD by 2--9\%.
Table~\ref{tab:speed} adds the SPEED-Bench throughput tiers of 8K to 32K tokens.
On Llama-3.1-8B, the speedup of DFlash falls from 2.84$\times$ at 8K tokens to 2.07$\times$ at 32K, while FOLD and ReRoPE stay between 3.81$\times$ and 4.02$\times$.
On Qwen3.5-9B and Gemma-4-12B, whose drafters degrade less over this range, the methods differ less.
ReRoPE is the fastest on Qwen3.5-9B and DFlash on Gemma-4-12B at 8K and 16K, while FOLD is the fastest on both at 32K, tied with ReRoPE on Gemma-4-12B.
CMR attains the highest $\tau$ on Qwen3.5-9B but a lower speedup, as on the other benchmarks.
With windows of 4,096 entries on these two drafters, a substantial fraction of the 8K and 16K inputs lies within the window, which leaves FOLD less to change.

\begin{table}[!htbp]
\centering
\footnotesize
\setlength{\tabcolsep}{2.7pt}
\renewcommand{\arraystretch}{0.95}\setlength{\aboverulesep}{0.3ex}\setlength{\belowrulesep}{0.5ex}
\begin{tabular*}{\columnwidth}{@{\extracolsep{\fill}}lcccccc@{}}
\toprule
\multirow{2}{*}{Method} & \multicolumn{2}{c}{8K} & \multicolumn{2}{c}{16K} & \multicolumn{2}{c}{32K} \\
\cmidrule(lr){2-3}\cmidrule(lr){4-5}\cmidrule(lr){6-7}
 & $\tau$ & Spd. & $\tau$ & Spd. & $\tau$ & Spd. \\
\midrule
\multicolumn{7}{l}{\textit{Llama-3.1-8B}} \\
DFlash & 3.02 & 2.84$\times$ & 2.69 & 2.52$\times$ & 2.21 & 2.07$\times$ \\
StreamingLLM & 3.50 & 3.27$\times$ & 3.18 & 2.99$\times$ & 2.89 & 2.72$\times$ \\
YaRN & 4.07 & 3.70$\times$ & 3.62 & 3.17$\times$ & 3.34 & 2.74$\times$ \\
CMR & 3.87 & 2.92$\times$ & 3.70 & 2.82$\times$ & 3.33 & 2.57$\times$ \\
ReRoPE & \textbf{4.31} & \textbf{4.02$\times$} & \textit{4.28} & \textbf{3.97$\times$} & \textit{4.13} & \textit{3.81$\times$} \\
InfLLM & 4.27 & 3.24$\times$ & 4.18 & \textit{3.21$\times$} & 3.97 & 3.09$\times$ \\
\textbf{FOLD (ours)} & \textit{4.30} & \textit{3.96$\times$} & \textbf{4.33} & \textbf{3.97$\times$} & \textbf{4.19} & \textbf{3.82$\times$} \\
\midrule
\multicolumn{7}{l}{\textit{Qwen3.5-9B}} \\
DFlash & 3.96 & 2.77$\times$ & 3.74 & 2.63$\times$ & 3.78 & 2.68$\times$ \\
StreamingLLM & 3.80 & 2.67$\times$ & 3.55 & 2.50$\times$ & 3.67 & 2.61$\times$ \\
YaRN & 3.93 & 2.75$\times$ & 3.76 & 2.65$\times$ & 3.80 & 2.70$\times$ \\
CMR & \textbf{4.56} & 2.83$\times$ & \textbf{4.61} & 2.89$\times$ & \textbf{4.64} & 2.89$\times$ \\
ReRoPE & \textit{4.50} & \textbf{3.16$\times$} & \textit{4.50} & \textbf{3.18$\times$} & \textit{4.41} & \textit{3.12$\times$} \\
InfLLM & 3.97 & 2.78$\times$ & 3.79 & 2.67$\times$ & 3.89 & 2.73$\times$ \\
\textbf{FOLD (ours)} & 4.47 & \textit{3.13$\times$} & 4.37 & \textit{3.08$\times$} & \textit{4.41} & \textbf{3.13$\times$} \\
\midrule
\multicolumn{7}{l}{\textit{Gemma-4-12B}} \\
DFlash & \textbf{3.31} & \textbf{3.05$\times$} & 3.32 & \textbf{3.04$\times$} & 2.95 & \textit{2.69$\times$} \\
StreamingLLM & 2.84 & 2.62$\times$ & 2.74 & 2.51$\times$ & 2.70 & 2.48$\times$ \\
YaRN & \textit{3.29} & \textit{3.00$\times$} & 3.32 & 2.96$\times$ & 3.00 & 2.60$\times$ \\
CMR & 3.09 & 2.52$\times$ & 3.10 & 2.52$\times$ & 2.97 & 2.41$\times$ \\
ReRoPE & \textit{3.29} & 2.98$\times$ & \textbf{3.35} & \textit{3.01$\times$} & \textbf{3.08} & \textbf{2.75$\times$} \\
InfLLM & \textit{3.29} & 2.59$\times$ & 3.29 & 2.58$\times$ & \textit{3.01} & 2.36$\times$ \\
\textbf{FOLD (ours)} & 3.27 & 2.97$\times$ & \textit{3.33} & 2.99$\times$ & \textbf{3.08} & \textbf{2.75$\times$} \\
\bottomrule
\end{tabular*}
\caption{\textbf{Results on SPEED-Bench.} Acceptance length $\tau$ and speedup over AR decoding on the throughput tiers of 8K, 16K, and 32K tokens. Bold and italics mark the best and second-best values per column and model.}
\label{tab:speed}
\end{table}

\subsection{Qwen3-8B with DFlash}
\label{app:qwen3_dflash}

Under the 32K context limit, we evaluate Qwen3-8B with DFlash on Gov and LBv2 (Table~\ref{tab:qwen3_dflash}).
FOLD achieves the highest speedup on Gov (2.63$\times$).
On LBv2, StreamingLLM outperforms FOLD (4.59$\times$ vs.\ 4.00$\times$), while FOLD's window-only variant reaches 4.61$\times$ with $\tau=5.25$.
The advantage of windowing is consistent with LBv2's short outputs (median 39 tokens) and prompt structure, which places the question, options, and answer format at the end.
On Gov, where summaries draw on the full report, far memory instead improves acceptance, raising $\tau$ from 2.51 to 2.85 over the window alone.

\begin{table}[t]
\centering
\footnotesize
\setlength{\tabcolsep}{2.7pt}
\renewcommand{\arraystretch}{0.95}\setlength{\aboverulesep}{0.3ex}\setlength{\belowrulesep}{0.5ex}
\begin{tabular*}{\columnwidth}{@{\extracolsep{\fill}}lcccc@{}}
\toprule
\multirow{2}{*}{Method} & \multicolumn{2}{c}{Gov} & \multicolumn{2}{c}{LBv2} \\
\cmidrule(lr){2-3}\cmidrule(lr){4-5}
 & $\tau$ & Spd. & $\tau$ & Spd. \\
\midrule
DFlash & 2.68 & 2.52$\times$ & 4.08 & 3.58$\times$ \\
StreamingLLM & 2.55 & 2.38$\times$ & 5.28 & \textbf{4.59$\times$} \\
YaRN & \textit{2.84} & 2.58$\times$ & \textbf{5.36} & \textit{4.26$\times$} \\
CMR & 2.77 & 2.12$\times$ & \textit{5.33} & 3.71$\times$ \\
ReRoPE & 2.80 & \textit{2.60$\times$} & 4.32 & 3.72$\times$ \\
InfLLM & 2.79 & 2.16$\times$ & 4.58 & 3.31$\times$ \\
\textbf{FOLD (ours)} & \textbf{2.85} & \textbf{2.63$\times$} & 4.72 & 4.00$\times$ \\
\bottomrule
\end{tabular*}
\caption{\textbf{Results for Qwen3-8B with its DFlash drafter.} Acceptance length $\tau$ and speedup over AR decoding. Bold and italics mark the best and second-best values per column.}
\label{tab:qwen3_dflash}
\end{table}

\subsection{Results in SGLang}
\label{app:sglang}

To check that the gains of FOLD are not specific to our vLLM implementation, we repeat the Llama-3.1-8B evaluation in a second engine, SGLang~0.5.18~\cite{zheng2024sglang}.

\paragraph{Setup.}
We use the same target model, the same DFlash drafter with block size $B=10$, and the 1,531 Llama-3.1-8B requests of the general tasks in Table~\ref{tab:main}, with the same token ids, generation limits, and stop tokens.
The run settings follow those of the main experiments: BF16, one request at a time, chunked prefill of 8,192 tokens, no prefix caching, CUDA graphs for the target and the drafter, and greedy decoding.
FOLD and the baselines are implemented in the SGLang drafter following their vLLM counterparts, and FOLD uses $W=2{,}048$ and $\alpha=2$.
We also report SGLang's built-in draft window, which restricts all block positions to one window of 2,048 entries, and Window, which keeps the window of FOLD without its far memory.
YaRN is omitted, as its SGLang implementation did not pass our numerical consistency checks.
Every method is paired with an AR run on the same GPU, and the speedup is computed as in the main experiments.

\paragraph{Output agreement.}
As in vLLM (Appendix~\ref{app:stats}), numerical differences between the drafting and AR paths can change the output when the two most likely tokens are nearly tied.
In a diagnostic run on a smaller set of requests, every difference we examined began at such a near tie.
In the full evaluation, every method, including the original drafter, differs from AR on 79 to 84 of the 1,531 requests, largely the same requests across methods (Table~\ref{tab:sglang}).
This exceeds the limit of about 2\% in our original protocol, so after observing these results we report all requests without examining each difference, and the numbers compare decoding on the same inputs rather than on identical outputs.
Removing the 94 requests that differ for any method changes every average by less than 0.01 and leaves the order of the methods unchanged.

\paragraph{Results.}
FOLD attains the highest average $\tau$ (3.26) and speedup (2.50$\times$, against 1.54$\times$ for the original drafter), with ReRoPE close behind at 2.46$\times$ (Table~\ref{tab:sglang}).

\begin{table}[!htbp]
\centering
\footnotesize
\setlength{\tabcolsep}{2.5pt}
\resizebox{\columnwidth}{!}{%
\begin{tabular}{lccccc}
\toprule
 & \multicolumn{3}{c}{All 1,531 requests} & \multicolumn{2}{c}{Common set (1,437)} \\
\cmidrule(lr){2-4}\cmidrule(lr){5-6}
Method & $\tau$ & Speedup & Diff.\ (AR) & $\tau$ & Speedup \\
\midrule
DFlash & 1.96 & 1.54$\times$ & 83 & 1.97 & 1.55$\times$ \\
SGLang window & 2.70 & 2.18$\times$ & 82 & 2.70 & 2.18$\times$ \\
StreamingLLM & 2.74 & 2.11$\times$ & 84 & 2.75 & 2.11$\times$ \\
CMR & 2.93 & 2.16$\times$ & 82 & 2.93 & 2.16$\times$ \\
ReRoPE & 3.21 & 2.46$\times$ & 79 & 3.22 & 2.47$\times$ \\
InfLLM & 3.15 & 2.15$\times$ & 80 & 3.15 & 2.15$\times$ \\
Window & 2.70 & 2.20$\times$ & 83 & 2.70 & 2.20$\times$ \\
\textbf{FOLD (ours)} & \textbf{3.26} & \textbf{2.50}$\times$ & 83 & \textbf{3.26} & \textbf{2.50}$\times$ \\
\bottomrule
\end{tabular}%
}
\caption{\textbf{Llama-3.1-8B in SGLang.} Averages over the seven general tasks with equal weight, with the speedup measured against AR on the same GPU. The Diff.\ (AR) column counts requests whose outputs differ from AR. The common set excludes the 94 requests that differ for any method. Bold indicates the highest acceptance length and speedup.}
\label{tab:sglang}
\end{table}

\section{Additional Analysis}
\label{app:analysis}

\subsection{Window and Memory by Input Length}
\label{app:decomp}

Table~\ref{tab:decomp_lbv2} extends Table~\ref{tab:decomp} to LBv2, whose requests span 10K to 250K tokens within one task.
The effect of the window becomes more favorable as inputs grow.
On Llama-3.1-8B, the change from DFlash to Window increases from $+0.57$ in the shortest bucket to $+1.34$ at 64--128K tokens, and on Qwen3.5-9B and Gemma-4-12B, it is negative or not detected below 128K tokens but positive beyond 128K ($+0.28$ for both).
The effect of the memory depends on the drafter.
It is positive in every bucket on Gemma-4-12B ($+0.33$ to $+0.56$) and largest beyond 128K tokens on Qwen3.5-9B ($+0.42$), whereas on Llama-3.1-8B it decreases with length and is no longer detected at 64--128K tokens, where the window already accounts for most of the gain.
On Qwen3-8B, whose LBv2 inputs are at most 32K tokens, the memory lowers $\tau$, consistent with the short answers of this task (Appendix~\ref{app:qwen3_dflash}).

\begin{table}[!htbp]
\centering
\footnotesize
\resizebox{\columnwidth}{!}{
\begin{tabular}{lrcccc}
\toprule
Bucket & $n$ & $\tau_{\mathrm{DFlash}}$ & $\tau_{\mathrm{FOLD}}$ & DFlash $\rightarrow$ Window & Window $\rightarrow$ FOLD \\
\midrule
\multicolumn{6}{l}{\textit{Llama-3.1-8B}} \\
8--16K & 28 & 2.63 & 4.07 & +0.57 {\scriptsize[+0.36, +0.79]} & +0.87 {\scriptsize[+0.52, +1.25]} \\
16--32K & 90 & 2.85 & 3.99 & +0.66 {\scriptsize[+0.54, +0.78]} & +0.48 {\scriptsize[+0.32, +0.64]} \\
32--64K & 72 & 3.12 & 4.44 & +1.09 {\scriptsize[+0.91, +1.28]} & +0.23 {\scriptsize[+0.06, +0.43]} \\
64--128K & 130 & 2.83 & 4.18 & +1.34 {\scriptsize[+1.16, +1.52]} & +0.01 {\scriptsize[-0.08, +0.09]} \\
\midrule
\multicolumn{6}{l}{\textit{Qwen3-8B}} \\
8--16K & 16 & 4.81 & 5.08 & +0.86 {\scriptsize[+0.39, +1.35]} & -0.60 {\scriptsize[-1.12, -0.08]} \\
16--32K & 63 & 3.89 & 4.62 & +1.26 {\scriptsize[+1.01, +1.49]} & -0.52 {\scriptsize[-0.72, -0.32]} \\
\midrule
\multicolumn{6}{l}{\textit{Qwen3.5-9B}} \\
8--16K & 20 & 4.42 & 4.37 & -0.06 {\scriptsize[-0.22, +0.05]} & +0.00 {\scriptsize[+0.00, +0.01]} \\
16--32K & 85 & 4.51 & 4.61 & -0.19 {\scriptsize[-0.37, -0.01]} & +0.30 {\scriptsize[+0.15, +0.48]} \\
32--64K & 66 & 4.38 & 4.55 & +0.14 {\scriptsize[-0.08, +0.40]} & +0.03 {\scriptsize[-0.20, +0.22]} \\
64--128K & 107 & 4.24 & 4.60 & +0.05 {\scriptsize[-0.08, +0.21]} & +0.31 {\scriptsize[+0.08, +0.55]} \\
$\geq$128K & 101 & 4.05 & 4.75 & +0.28 {\scriptsize[+0.09, +0.52]} & +0.42 {\scriptsize[+0.22, +0.64]} \\
\midrule
\multicolumn{6}{l}{\textit{Gemma-4-12B}} \\
8--16K & 22 & 3.40 & 3.33 & -0.56 {\scriptsize[-0.72, -0.39]} & +0.48 {\scriptsize[+0.29, +0.65]} \\
16--32K & 83 & 3.55 & 3.54 & -0.57 {\scriptsize[-0.69, -0.46]} & +0.56 {\scriptsize[+0.47, +0.66]} \\
32--64K & 72 & 3.20 & 3.44 & -0.31 {\scriptsize[-0.42, -0.21]} & +0.55 {\scriptsize[+0.46, +0.66]} \\
64--128K & 102 & 2.38 & 2.72 & -0.01 {\scriptsize[-0.09, +0.08]} & +0.35 {\scriptsize[+0.26, +0.42]} \\
$\geq$128K & 99 & 2.25 & 2.86 & +0.28 {\scriptsize[+0.22, +0.35]} & +0.33 {\scriptsize[+0.26, +0.39]} \\
\bottomrule
\end{tabular}
}
\caption{\textbf{Decomposition of the difference in $\tau$ on LBv2 by input length.} The last two columns are changes in $\tau$ paired by request, which sum to the difference between FOLD and DFlash, with pointwise 95\% bootstrap intervals.}
\label{tab:decomp_lbv2}
\end{table}

\subsection{Concentration Factor}
\label{app:alpha}

\begin{table}[t]
\centering
\footnotesize
\setlength{\tabcolsep}{3pt}
\renewcommand{\arraystretch}{0.95}\setlength{\aboverulesep}{0.3ex}\setlength{\belowrulesep}{0.5ex}
\resizebox{\columnwidth}{!}{%
\begin{tabular}{@{}llrccc@{}}
\toprule
Model & Tasks & $n$ & $\tau_{\alpha=2}$ & $\alpha=2$ vs.\ 1 & $\alpha=4$ vs.\ 2 \\
\midrule
Llama & Natural & 1137 & 3.15 & 1.174 {\scriptsize[1.158, 1.191]} & 1.027$^\dagger$ {\scriptsize[1.022, 1.032]} \\
 & SPEED & 1816 & 4.24 & 1.039 {\scriptsize[1.036, 1.043]} & 0.991 {\scriptsize[0.989, 0.992]} \\
 & QA/MC/Dia & 391 & 3.22 & 1.078 {\scriptsize[1.070, 1.086]} & 0.993 {\scriptsize[0.991, 0.996]} \\
\midrule
Qwen3 & Natural & 359 & 3.26 & 1.160 {\scriptsize[1.138, 1.183]} & 1.012$^\dagger$ {\scriptsize[1.006, 1.019]} \\
 & SPEED & 1150 & 3.51 & 1.194 {\scriptsize[1.185, 1.203]} & 1.012 {\scriptsize[1.010, 1.014]} \\
\midrule
Qwen3.5 & Natural & 1138 & 3.75 & 2.709 {\scriptsize[2.632, 2.787]} & 1.217 {\scriptsize[1.198, 1.236]} \\
 & SPEED & 1771 & 4.40 & 1.545 {\scriptsize[1.522, 1.568]} & 0.994 {\scriptsize[0.993, 0.995]} \\
 & QA/MC/Dia & 662 & 3.32 & 2.938 {\scriptsize[2.910, 2.967]} & 1.022 {\scriptsize[1.020, 1.025]} \\
\midrule
Gemma & Natural & 1134 & 3.13 & 1.070 {\scriptsize[1.063, 1.076]} & 1.002$^\dagger$ {\scriptsize[0.998, 1.006]} \\
 & SPEED & 1775 & 3.21 & 1.004 {\scriptsize[1.002, 1.006]} & 0.998 {\scriptsize[0.996, 0.9995]} \\
 & QA/MC/Dia & 662 & 2.47 & 1.074 {\scriptsize[1.067, 1.083]} & 0.999 {\scriptsize[0.994, 1.004]} \\
\bottomrule
\end{tabular}%
}
\caption{\textbf{Effect of $\alpha$ on the general tasks and SPEED-Bench.} Ratios and $^\dagger$ as in Table~\ref{tab:alpha}, with $n$ the number of requests. Natural covers Gov, LBv2, Retr.KV, and En.Sum, QA/MC/Dia covers En.QA, En.MC, and En.Dia, and SPEED covers all SPEED-Bench requests of each model that are longer than the window.}
\label{tab:alpha_full}
\end{table}

Table~\ref{tab:alpha_full} splits the general tasks of Table~\ref{tab:alpha} into Natural and QA/MC/Dia and adds SPEED-Bench.
Raising $\alpha$ from 1 to 2 also yields mean ratios above 1 on SPEED-Bench for every model.
On Llama-3.1-8B and Qwen3.5-9B, this gain is smaller than on their Natural tasks (mean ratios of 1.039 vs.\ 1.174 and 1.545 vs.\ 2.709), consistent with the fewer distant entries of inputs of at most 32K tokens.
Raising $\alpha$ further to 4 keeps the mean ratio on SPEED-Bench within 1.2\% of 1.
For the cells marked $\dagger$, the ratio of the overall $\tau$ of all requests, computed from their total accepted tokens and verification rounds, is 0.995 [0.993, 0.997] for Llama-3.1-8B, 0.999 [0.997, 1.001] for Qwen3-8B, and 0.996 [0.992, 0.9998] for Gemma-4-12B in Table~\ref{tab:alpha}, and 0.997 [0.995, 0.9997] for Llama-3.1-8B and 0.994 [0.988, 0.999] for Gemma-4-12B on the Natural tasks in Table~\ref{tab:alpha_full}.

%% file: custom.bib
@inproceedings{leviathan2023fast,
  title={Fast inference from transformers via speculative decoding},
  author={Leviathan, Yaniv and Kalman, Matan and Matias, Yossi},
  booktitle={International conference on machine learning},
  pages={19274--19286},
  year={2023},
  organization={PMLR}
}

@article{chen2023accelerating,
  title={Accelerating large language model decoding with speculative sampling},
  author={Chen, Charlie and Borgeaud, Sebastian and Irving, Geoffrey and Lespiau, Jean-Baptiste and Sifre, Laurent and Jumper, John},
  journal={arXiv preprint arXiv:2302.01318},
  year={2023}
}

@article{li2025eagle3,
  title={{EAGLE-3}: Scaling up inference acceleration of large language models via training-time test},
  author={Li, Yuhui and Wei, Fangyun and Zhang, Chao and Zhang, Hongyang},
  journal={Advances in Neural Information Processing Systems},
  volume={38},
  pages={136737--136756},
  year={2025}
}

@inproceedings{chen2026dflash,
  title={{DFlash}: Block diffusion for flash speculative decoding},
  author={Chen, Jian and Liang, Yesheng and Liu, Zhijian},
  booktitle={International Conference on Machine Learning},
  pages={15343--15355},
  year={2026},
  organization={PMLR}
}

@article{huang2026domino,
  title={Domino: Decoupling causal modeling from autoregressive drafting in speculative decoding},
  author={Huang, Jianuo and Zhang, Yaojie and Zhang, Qituan and Lin, Hao and Xu, Hanlin and Zhang, Linfeng},
  journal={arXiv preprint arXiv:2605.29707},
  year={2026}
}

@article{cheng2026dspark,
  title={{DSpark}: Confidence-Scheduled Speculative Decoding with Semi-Autoregressive Generation},
  author={Cheng, Xin and Yu, Xingkai and Shao, Chenze and Li, Jiashi and Xiong, Yunfan and Qian, Yi and Zhu, Jiaqi and Ma, Shirong and Zhang, Xiaokang and Ye, Jiasheng and others},
  journal={arXiv preprint arXiv:2607.05147},
  year={2026}
}

@inproceedings{sadhukhan2025magicdec,
  title={{MagicDec}: Breaking the latency-throughput tradeoff for long context generation with speculative decoding},
  author={Sadhukhan, Ranajoy and Chen, Jian and Chen, Zhuoming and Tiwari, Vashisth and Lai, Ruihang and Shi, Jinyuan and Yen, Ian and May, Avner and Chen, Tianqi and Chen, Beidi},
  booktitle={International Conference on Learning Representations},
  volume={2025},
  pages={6835--6850},
  year={2025}
}

@inproceedings{cha2026specextend,
  title={{SpecExtend}: A drop-in enhancement for speculative decoding of long sequences},
  author={Cha, Jungyoub and Kim, Hyunjong and Cho, Sungzoon},
  booktitle={Findings of the Association for Computational Linguistics: ACL 2026},
  pages={43366--43377},
  year={2026}
}

@inproceedings{sun2024triforce,
  title={{TriForce}: Lossless acceleration of long sequence generation with hierarchical speculative decoding},
  author={Sun, Hanshi and Chen, Zhuoming and Yang, Xinyu and Tian, Yuandong and Chen, Beidi},
  booktitle={First Conference on Language Modeling},
  year={2024}
}

@article{jiang2026hspec,
  title={{H-Spec}: Parallel Speculative Decoding Without a Drafter-Side {KV} Cache},
  author={Jiang, Weifan and Chitty-Venkata, Krishna Teja and Flynn, Megan and Meyerson, Reed and Qi, Zhenting and Wu, Tianyu and Kurtic, Eldar and Yu, Minlan and Marques, Alexandre},
  journal={arXiv preprint arXiv:2609.24197},
  year={2026}
}

@inproceedings{yang2026longspec,
  title={{LongSpec}: Long-Context Lossless Speculative Decoding with Efficient Drafting and Verification},
  author={Yang, Penghui and Du, Cunxiao and Zhang, Fengzhuo and Wang, Haonan and Pang, Tianyu and Du, Chao and An, Bo},
  booktitle={Proceedings of the 64th Annual Meeting of the Association for Computational Linguistics (Volume 1: Long Papers)},
  pages={1826--1844},
  year={2026}
}

@inproceedings{abramovich2026speedbench,
  title={{SPEED-Bench}: A unified and diverse benchmark for speculative decoding},
  author={Abramovich, Talor and Ashkenazi, Maor and Putterman, Izzy and Chislett, Benjamin and Mitra, Tiyasa and Darvish Rouhani, Bita and Zilberstein, Ran and Geifman, Yonatan},
  booktitle={International Conference on Machine Learning},
  pages={240--267},
  year={2026},
  organization={PMLR}
}

@article{rusanovsky2026lilicorr,
  title={{LiLiCorr}: Lightweight Likelihood Correlation of Parallel Drafts for Speculative Decoding},
  author={Rusanovsky, Matan and Miron, Yoav and Uziel, Roy and Belhasin, Omer and Zilberstein, Ran and Ashkenazi, Maor and Elad, Michael},
  journal={arXiv preprint arXiv:2608.20530},
  year={2026}
}

@article{yue2026vegas,
  title={Vegas: Self-Speculative Decoding with Verification-Guided Sparse Attention},
  author={Yue, Yikang and Xue, Yuqi and Huang, Jian},
  journal={arXiv preprint arXiv:2602.07223},
  year={2026}
}

@article{chen2023extending,
  title={Extending context window of large language models via positional interpolation},
  author={Chen, Shouyuan and Wong, Sherman and Chen, Liangjian and Tian, Yuandong},
  journal={arXiv preprint arXiv:2306.15595},
  year={2023}
}

@inproceedings{peng2024yarn,
  title={{YaRN}: Efficient context window extension of large language models},
  author={Peng, Bowen and Quesnelle, Jeffrey and Fan, Honglu and Shippole, Enrico},
  booktitle={International Conference on Learning Representations},
  volume={2024},
  pages={31932--31951},
  year={2024}
}

@misc{su2023rerope,
  title        = {Rectified Rotary Position Embeddings},
  author       = {Su, Jianlin},
  year         = {2023},
  howpublished = {\url{https://github.com/bojone/rerope}},
  url          = {https://github.com/bojone/rerope}
}

@inproceedings{han2024lminfinite,
  title={{LM-Infinite}: Zero-shot extreme length generalization for large language models},
  author={Han, Chi and Wang, Qifan and Peng, Hao and Xiong, Wenhan and Chen, Yu and Ji, Heng and Wang, Sinong},
  booktitle={Proceedings of the 2024 Conference of the North American Chapter of the Association for Computational Linguistics: Human Language Technologies (Volume 1: Long Papers)},
  pages={3991--4008},
  year={2024}
}

@article{jin2024llm,
  title={{LLM} Maybe {LongLM}: {Self-Extend} {LLM} Context Window Without Tuning},
  author={Jin, Hongye and Han, Xiaotian and Yang, Jingfeng and Jiang, Zhimeng and Liu, Zirui and Chang, Chia-Yuan and Chen, Huiyuan and Hu, Xia},
  journal={arXiv preprint arXiv:2401.01325},
  year={2024}
}

@article{an2024training,
  title={Training-free long-context scaling of large language models},
  author={An, Chenxin and Huang, Fei and Zhang, Jun and Gong, Shansan and Qiu, Xipeng and Zhou, Chang and Kong, Lingpeng},
  journal={arXiv preprint arXiv:2402.17463},
  year={2024}
}

@inproceedings{an2025string,
  title={Why does the effective context length of {LLMs} fall short?},
  author={An, Chenxin and Zhang, Jun and Zhong, Ming and Li, Lei and Gong, Shansan and Luo, Yao and Xu, Jingjing and Kong, Lingpeng},
  booktitle={International Conference on Learning Representations},
  volume={2025},
  pages={54791--54810},
  year={2025}
}

@article{tang2026jetlong,
  title={{Jet-Long}: Efficient Long-Context Extension with Dynamic Bifocal {RoPE}},
  author={Tang, Haozhan and Wang, Zerui and Gu, Yuxian and Han, Song and Cai, Han},
  journal={arXiv preprint arXiv:2607.07740},
  year={2026}
}

@inproceedings{xiao2024streamingllm,
  title={Efficient streaming language models with attention sinks},
  author={Xiao, Guangxuan and Tian, Yuandong and Chen, Beidi and Han, Song and Lewis, Mike},
  booktitle={International Conference on Learning Representations},
  volume={2024},
  pages={21875--21895},
  year={2024}
}

@article{xiao2024infllm,
  title={{InfLLM}: Training-free long-context extrapolation for {LLMs} with an efficient context memory},
  author={Xiao, Chaojun and Zhang, Pengle and Han, Xu and Xiao, Guangxuan and Lin, Yankai and Zhang, Zhengyan and Liu, Zhiyuan and Sun, Maosong},
  journal={Advances in neural information processing systems},
  volume={37},
  pages={119638--119661},
  year={2024}
}

@article{zhu2026tracelab,
  title={{TraceLab}: Characterizing Coding Agent Workloads for {LLM} Serving},
  author={Zhu, Kan and Jacob, Mathew and Ma, Chenxi and Pan, Yi and Wang, Stephanie and Krishnamurthy, Arvind and Kasikci, Baris},
  journal={arXiv preprint arXiv:2606.30560},
  year={2026}
}

@article{su2024roformer,
  title={{RoFormer}: Enhanced transformer with rotary position embedding},
  author={Su, Jianlin and Ahmed, Murtadha and Lu, Yu and Pan, Shengfeng and Bo, Wen and Liu, Yunfeng},
  journal={Neurocomputing},
  volume={568},
  pages={127063},
  year={2024},
  publisher={Elsevier}
}

@inproceedings{haviv2022transformer,
  title={Transformer language models without positional encodings still learn positional information},
  author={Haviv, Adi and Ram, Ori and Press, Ofir and Izsak, Peter and Levy, Omer},
  booktitle={Findings of the Association for Computational Linguistics: EMNLP 2022},
  pages={1382--1390},
  year={2022}
}

@article{grattafiori2024llama3,
  title={The {Llama} 3 herd of models},
  author={Grattafiori, Aaron and Dubey, Abhimanyu and Jauhri, Abhinav and Pandey, Abhinav and Kadian, Abhishek and Al-Dahle, Ahmad and Letman, Aiesha and Mathur, Akhil and Schelten, Alan and Vaughan, Alex and others},
  journal={arXiv preprint arXiv:2407.21783},
  year={2024}
}

@inproceedings{bai2024longbench,
  title={{LongBench}: A bilingual, multitask benchmark for long context understanding},
  author={Bai, Yushi and Lv, Xin and Zhang, Jiajie and Lyu, Hongchang and Tang, Jiankai and Huang, Zhidian and Du, Zhengxiao and Liu, Xiao and Zeng, Aohan and Hou, Lei and others},
  booktitle={Proceedings of the 62nd annual meeting of the association for computational linguistics (volume 1: Long papers)},
  pages={3119--3137},
  year={2024}
}

@inproceedings{bai2025longbenchv2,
  title={{LongBench} v2: Towards deeper understanding and reasoning on realistic long-context multitasks},
  author={Bai, Yushi and Tu, Shangqing and Zhang, Jiajie and Peng, Hao and Wang, Xiaozhi and Lv, Xin and Cao, Shulin and Xu, Jiazheng and Hou, Lei and Dong, Yuxiao and others},
  booktitle={Proceedings of the 63rd Annual Meeting of the Association for Computational Linguistics (Volume 1: Long Papers)},
  pages={3639--3664},
  year={2025}
}

@inproceedings{zhang2024infinitebench,
  title={{$\infty$Bench}: Extending long context evaluation beyond {100K} tokens},
  author={Zhang, Xinrong and Chen, Yingfa and Hu, Shengding and Xu, Zihang and Chen, Junhao and Hao, Moo and Han, Xu and Thai, Zhen and Wang, Shuo and Liu, Zhiyuan and others},
  booktitle={Proceedings of the 62nd Annual Meeting of the Association for Computational Linguistics (Volume 1: Long Papers)},
  pages={15262--15277},
  year={2024}
}

@inproceedings{kwon2023vllm,
  title={Efficient memory management for large language model serving with {PagedAttention}},
  author={Kwon, Woosuk and Li, Zhuohan and Zhuang, Siyuan and Sheng, Ying and Zheng, Lianmin and Yu, Cody Hao and Gonzalez, Joseph and Zhang, Hao and Stoica, Ion},
  booktitle={Proceedings of the 29th symposium on operating systems principles},
  pages={611--626},
  year={2023}
}

@misc{qwen35,
  title        = {{Qwen3.5-9B}},
  author       = {{Qwen Team}},
  year         = {2026},
  howpublished = {\url{https://huggingface.co/Qwen/Qwen3.5-9B}},
  note         = {Model release}
}

@misc{gemma4,
  title        = {{Gemma 4}},
  author       = {{Google DeepMind}},
  year         = {2026},
  howpublished = {\url{https://huggingface.co/google/gemma-4-12B-it}},
  note         = {Model release}
}

@misc{deepspec,
  title        = {{DeepSpec}: A Full-Stack Codebase for Training and Evaluating Draft Models for Speculative Decoding},
  author       = {{DeepSeek-AI}},
  year         = {2026},
  howpublished = {\url{https://github.com/deepseek-ai/DeepSpec}}
}

@article{yang2025qwen3,
  title={Qwen3 technical report},
  author={Yang, An and Li, Anfeng and Yang, Baosong and Zhang, Beichen and Hui, Binyuan and Zheng, Bo and Yu, Bowen and Gao, Chang and Huang, Chengen and Lv, Chenxu and others},
  journal={arXiv preprint arXiv:2505.09388},
  year={2025}
}

@article{zheng2024sglang,
  title={{SGLang}: Efficient execution of structured language model programs},
  author={Zheng, Lianmin and Yin, Liangsheng and Xie, Zhiqiang and Sun, Chuyue and Huang, Jeff and Yu, Cody H and Cao, Shiyi and Kozyrakis, Christos and Stoica, Ion and Gonzalez, Joseph E and others},
  journal={Advances in neural information processing systems},
  volume={37},
  pages={62557--62583},
  year={2024}
}

@article{dao2022flashattention,
  title={{FlashAttention}: Fast and memory-efficient exact attention with {IO}-awareness},
  author={Dao, Tri and Fu, Dan and Ermon, Stefano and Rudra, Atri and R{\'e}, Christopher},
  journal={Advances in neural information processing systems},
  volume={35},
  pages={16344--16359},
  year={2022}
}

@inproceedings{tillet2019triton,
  title={Triton: an intermediate language and compiler for tiled neural network computations},
  author={Tillet, Philippe and Kung, Hsiang-Tsung and Cox, David},
  booktitle={Proceedings of the 3rd ACM SIGPLAN International Workshop on Machine Learning and Programming Languages},
  pages={10--19},
  year={2019}
}

@article{paszke2019pytorch,
  title={{PyTorch}: An imperative style, high-performance deep learning library},
  author={Paszke, Adam and Gross, Sam and Massa, Francisco and Lerer, Adam and Bradbury, James and Chanan, Gregory and Killeen, Trevor and Lin, Zeming and Gimelshein, Natalia and Antiga, Luca and others},
  journal={Advances in neural information processing systems},
  volume={32},
  year={2019}
}

@inproceedings{liu2024repobench,
  title={Repobench: Benchmarking repository-level code auto-completion systems},
  author={Liu, Tianyang and Xu, Canwen and McAuley, Julian},
  booktitle={International Conference on Learning Representations},
  volume={2024},
  pages={47832--47850},
  year={2024}
}

@article{bogomolov2024long,
  title={Long code arena: a set of benchmarks for long-context code models},
  author={Bogomolov, Egor and Eliseeva, Aleksandra and Galimzyanov, Timur and Glukhov, Evgeniy and Shapkin, Anton and Tigina, Maria and Golubev, Yaroslav and Kovrigin, Alexander and Van Deursen, Arie and Izadi, Maliheh and others},
  journal={arXiv preprint arXiv:2406.11612},
  year={2024}
}

@inproceedings{jimenez2024swebench,
  title={Swe-bench: Can language models resolve real-world github issues?},
  author={Jimenez, Carlos E and Yang, John and Wettig, Alexander and Yao, Shunyu and Pei, Kexin and Press, Ofir and Narasimhan, Karthik},
  booktitle={International Conference on Learning Representations},
  volume={2024},
  pages={54107--54157},
  year={2024}
}
